\RequirePackage{fix-cm}
\documentclass[smallextended]{svjour3}       % onecolumn (second format)
\smartqed  % flush right qed marks, e.g. at end of proof
\usepackage{graphicx}
\usepackage{tcolorbox}
\usepackage{hyperref} 
\usepackage{booktabs}
\usepackage{array}
\usepackage{subcaption}
\usepackage[accsupp]{axessibility}  % Improves PDF readability for those with disabilities.
\usepackage{multirow}
\usepackage{authblk}
\journalname{Optimization and Engineering}
\begin{document}

\title{A Robust Anchor-based Method for Multi-Camera Pedestrian Localization} %\thanks{Grants or other notes

\subtitle{}


\author{Wanyu Zhang \and
Jiaqi Zhang \and
Dongdong Ge \and
Yu Lin \and
Huiwen Yang \and 
Huikang Liu \and 
Yinyu Ye
}

\authorrunning{W.~Zhang et al.} % if too long for running head

\institute{
    Wanyu Zhang \and Jiaqi Zhang \at
    Shanghai University of Finance and Economics, China \\
    \email{wanyuzhang@stu.sufe.edu.cn}, \email{johnzhang@163.sufe.edu.cn}
    \and
    Dongdong Ge \and Huikang Liu \at
    Shanghai Jiao Tong University, China \\
    \email{ddge@sjtu.edu.cn}, \email{hkl1u@sjtu.edu.cn}
    \and
    Yu Lin \and Huiwen Yang \at
    CUE Group, China \\
    \email{yu.lin@cue.group}, \email{yang.huiwen@cue.group}
    \and
    Yinyu Ye \at
    Stanford University, United States \\
    \email{yyye@stanford.edu}
}

\date{Received: date / Accepted: date}

\maketitle

\begin{abstract}
This paper addresses the problem of vision-based pedestrian localization, which estimates a pedestrian's location using images and camera parameters. In practice, however, calibrated camera parameters often deviate from the ground truth, leading to inaccuracies in localization. To address this issue, we propose an anchor-based method that leverages fixed-position anchors to reduce the impact of camera parameter errors. We provide a theoretical analysis that demonstrates the robustness of our approach. Experiments conducted on simulated, real-world, and public datasets show that our method significantly improves localization accuracy and remains resilient to noise in camera parameters, compared to methods without anchors.
\end{abstract}

\section{Introduction}
\label{sec:intro}

Location-based services have attracted considerable attention due to their critical applications across various industries, such as healthcare and smart commerce \cite{guo2023multi,pervs2011fusion}. In healthcare, these services play a vital role in monitoring patients and staff, managing equipment and inventory, and optimizing workflows. Similarly, in the retail sector, they enhance the shopping experience by analyzing customer behavior. Furthermore, the security and safety sectors can leverage this technology to effectively detect and respond to emergencies.

These important location-based applications have necessitated the need for accessible and precise localization technologies. Localization methods can be broadly categorized into tag-based and tag-free methods, depending on whether objects are required to carry small devices or sensors. In tag-based methods, objects equipped with tags interact with fixed infrastructures to determine their location. Below, we briefly review several tag-based methods and their localization precision:
\begin{itemize}
    \item Bluetooth localization \cite{wang2015rssi,kriz2016improving}: Precision can vary between 1 to 100 meters, largely depending on environmental factors such as obstacles and signal interference.
    \item RFID-based localization \cite{zhou2009rfid,ni2011rfid,papapostolou2011rfid}: Typically offers better accuracy, around 1 to 5 meters, but can be negatively impacted by metal surfaces and liquids.
    \item Ultra-Wideband (UWB) localization \cite{ye2010high,zwirello2012uwb}: Known for its high accuracy, UWB can achieve precision of 0.1 to 0.5 meters under ideal conditions.
\end{itemize}

Tag-free, vision-based localization methods, on the contrary, rely solely on images captured by cameras without utilizing sensor information. Vision-based methods can be broadly divided into two types: foreground projection-based and detection-based methods. Foreground projection-based methods detect moving objects using background subtraction models and then project the resulting foreground masks onto one or more reference planes \cite{khan2008tracking,jiuqing2012multi}. These projections are analyzed to understand relationships between the projected blobs, addressing challenges like interconnected masks and occlusions \cite{khan2008tracking,sun2010robust,weiming2006principal}. Some approaches in this category model occlusions using statistical algorithms, such as Bayesian models or probability occupancy maps, making them more robust in occlusion-heavy environments \cite{fleuret2007multicamera,yang20183d}.

However, foreground segmentation methods are highly sensitive to lighting conditions and frequent occlusions \cite{jiuqing2012multi}. Specifically, when background pixels vary due to changing illumination, these algorithms struggle to differentiate them from foreground objects. Furthermore, in crowded scenes with frequent occlusions, the robustness of object detection algorithms is significantly compromised. Recent research has focused on improving foreground detection accuracy through deep-learning models \cite{chavdarova2017deep}. Nonetheless, pixel-level processing remains computationally intensive, limiting its scalability and integration into tracking frameworks that require real-time location information.

\begin{figure}
    \centering
    \includegraphics[width=0.8\linewidth]{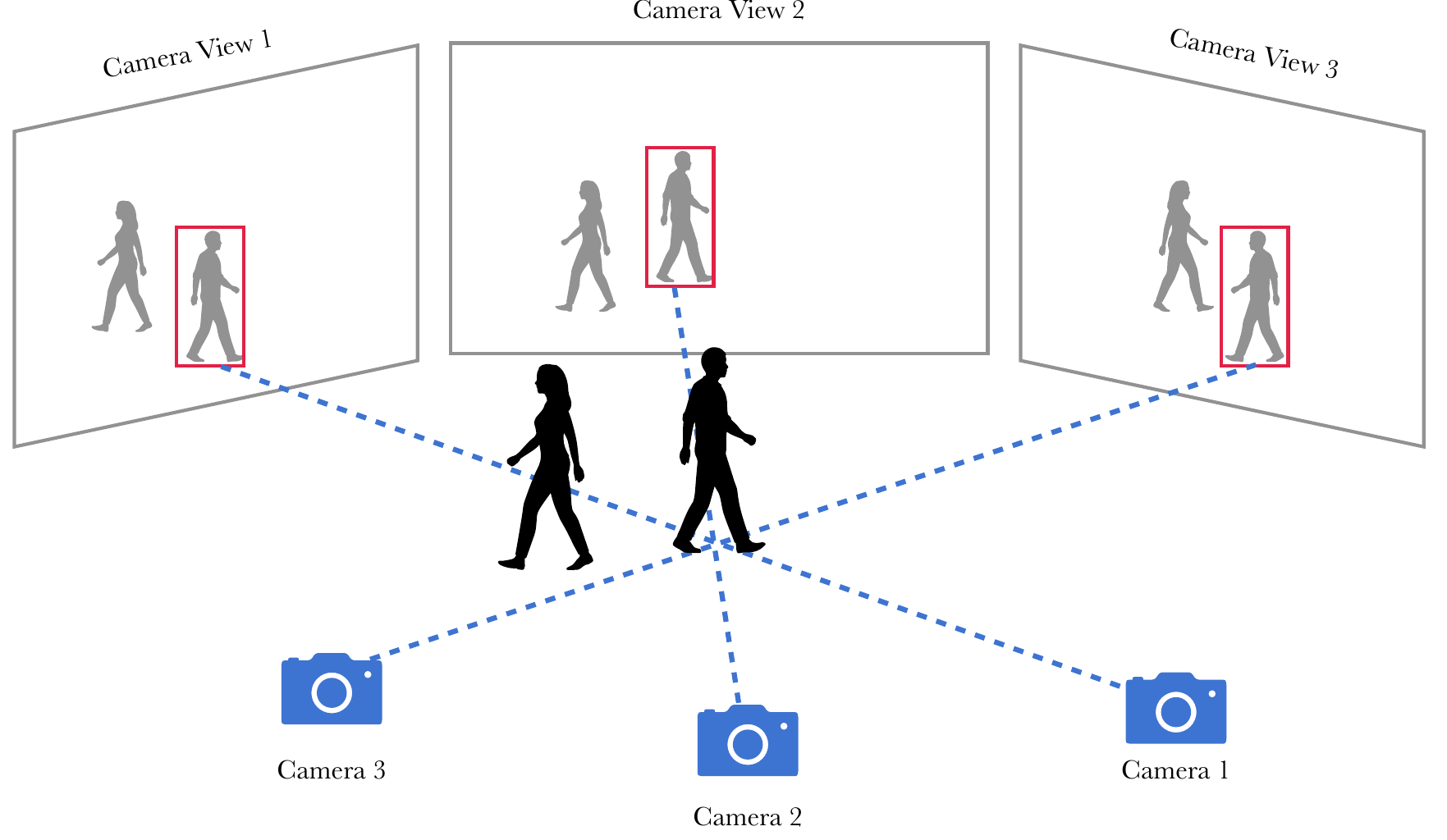}
    \caption{An illustration of detection-based localization problem. Given the input images (labeled by detection boxes) and parameters of multiple cameras, we aim to recover the location of the targets in space.}
    \label{fig:localization-graph}
\end{figure}

Detection-based localization methods, as illustrated in Figure \ref{fig:localization-graph}, preprocess camera images by labeling each person with a unique detection box and then calculate the object’s location in space. Essentially, this can be formulated as a reprojection error minimization task (see Figure \ref{fig: reprojection-illus} for illustration). This objective function is commonly used in computer vision tasks, such as bundle adjustment for 3D reconstruction \cite{agarwal2010bundle,triggs2000bundle} and camera calibration \cite{zhang2000flexible}. Specifically, for a hypothesized position, we reproject it onto the image plane using calibrated camera parameters. The reprojection error is defined as the distance between the reprojected pixel and the observed pixel in the image. When the camera parameters are accurate, this error should be minimized around the target’s true position.

Detection-based localization is also widely used in tracking algorithms to ensure spatial consistency. In \cite{huang2023enhancing}, the bottom center of the detection box, interpreted as the ankle point, is projected onto the ground plane to determine the grounding point, which is then used for ID reassignment. \cite{luna2022online} introduces an online clustering-based tracking method that calculates an object's location by averaging grounding points identified across multiple camera views. A more precise location estimation approach is proposed by \cite{kim2017deep}, where the ankle point is detected using a deep learning model, replacing the traditional use of the bottom center of the detection box.

\subsection{Outline and Contributions}

In this paper, we first observe that the effectiveness of the aforementioned detection-based localization methods relies heavily on the critical assumption that the provided camera parameters are accurate, ensuring that the reprojection process closely aligns with the ground-truth imaging process. While existing camera calibration techniques can achieve low reprojection errors in controlled environments \cite{moreno2012simple,zhang2000flexible}, real-world applications often expose significant deviations in camera parameters from the ground truth due to several factors, as outlined below:
\begin{itemize}
    \item[(i)] Calibration is typically optimized for the camera's central viewing area, but distortion effects become more pronounced in peripheral views.
    \item[(ii)]  External camera parameters are often estimated via manual measurements or 3D surveying, making precise determination costly and impractical when managing a large number of cameras.
    \item[(iii)] Over time, camera extrinsics can gradually shift due to environmental factors, such as long-term suspension or vibrations.
    \item[(iv)] In large-scale scenarios, precisely calibrating each camera is labor-intensive and impractical, leading to the adoption of sampling calibration.
\end{itemize}

In Section \ref{sec:no-anchor-model}, we show that, due to calibration errors and pixel extraction inaccuracies, directly minimizing reprojection error may prevent the optimal solution from converging to the ground-truth position. This observation leads us to establish the central research problem:
\begin{tcolorbox}
Given the potential inaccuracies in camera parameters, is it possible to achieve precise (centimeter-level) vision-based localization?
\end{tcolorbox}

In this paper, we give a positive answer and propose using anchors within the camera’s view to mitigate the impact of camera parameter errors. Anchors are points with known 3D positions, allowing us to observe the discrepancy between their reprojected and observed pixels. The reprojection error at these anchors reflects the camera parameter error. The underlying intuition is that, since both the localization targets and the anchors are affected by the same camera parameter errors, we can correct the reprojected pixels by "canceling" the camera parameter errors observed at the anchors.

\begin{figure}[htbp]
    \centering
    \begin{subfigure}[b]{\textwidth}
        \centering
        \includegraphics[width=\linewidth]{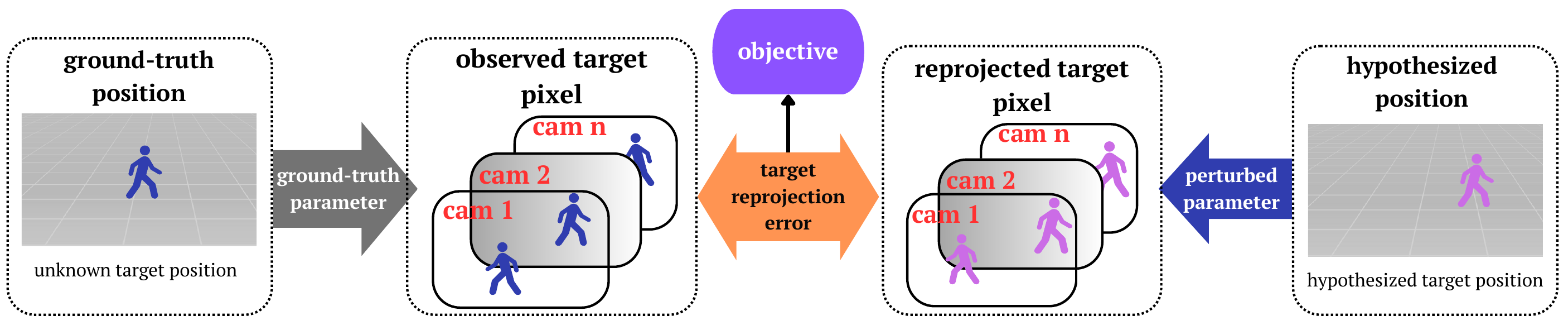}
        \caption{No-anchor localization method}
        \label{fig: reprojection-illus}
    \end{subfigure}
    \hfill
    \begin{subfigure}[b]{\textwidth}
        \centering
        \includegraphics[width=\linewidth]{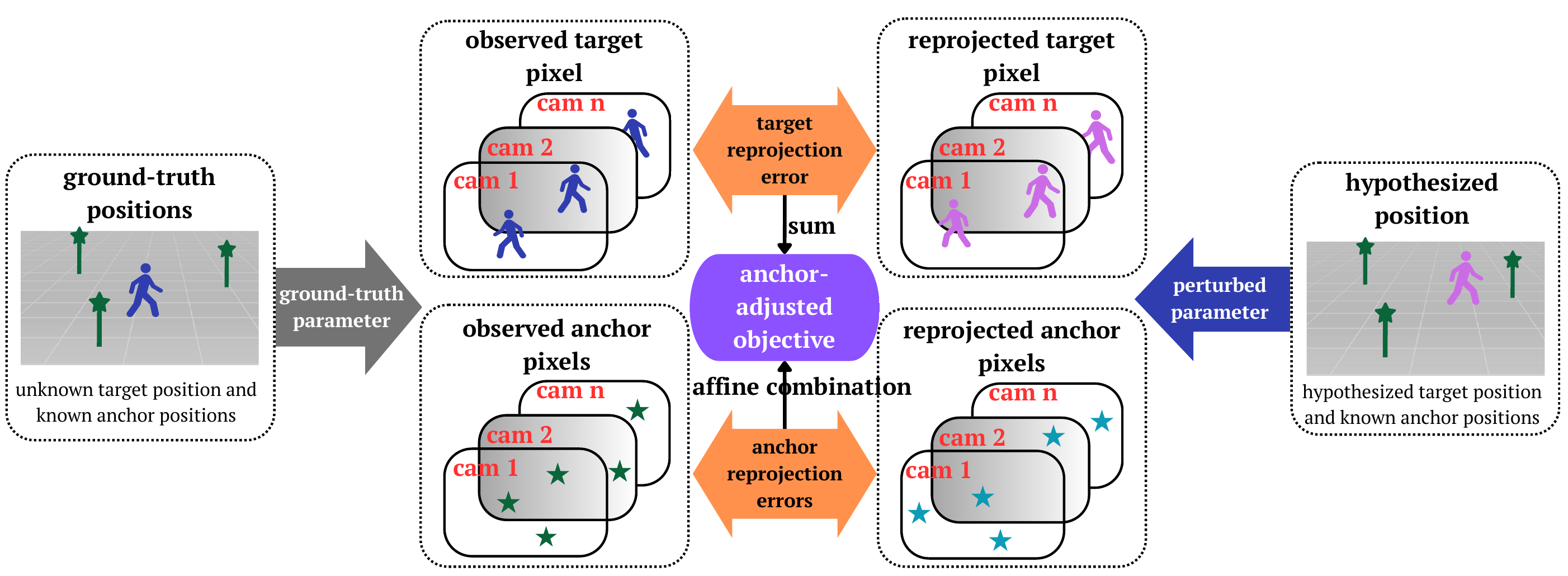}
        \caption{Anchor-based localization method}
        \label{fig: anchor-reprojection-illus}
    \end{subfigure}
    \caption{Comparison of no-anchor method and anchor-based method. The no-anchor localization method directly minimizes the reprojection error, whereas the anchor-based method minimizes the anchor-adjusted reprojection error.}
    \label{fig:localization-comparison}
\end{figure}

Through mathematical transformations, we show that camera parameter errors can be reduced by an affine combination of anchor reprojection errors. Based on this observation, we introduce an anchor-based localization method. Instead of directly minimizing reprojection errors, our method minimizes anchor-adjusted reprojection errors. An illustration of this approach is provided in Figure \ref{fig: anchor-reprojection-illus}. The proposed method improves localization precision and exhibits robustness against camera parameter errors.

Thus far, the localization task has been approached as a static problem, where the target's location in each frame is solved independently. However, temporal associations exist between frames, as the target's position in adjacent frames is unlikely to change drastically. To leverage this temporal information, we optimize the target's positions within a time window. In addition to minimizing the sum of anchor-adjusted reprojection errors over time, we introduce a smoothness penalty to reduce abrupt changes in the target's position between consecutive frames. Experiments on real-world data demonstrate that incorporating the smoothness penalty greatly enhances both the precision and stability of localization.

The main contributions of this paper are as follows:
\begin{itemize}
    \item We introduce a vision-based localization method that mitigates the impact of camera parameter errors by utilizing anchors.
    \item We provide a theoretical analysis of how camera parameter errors affect localization precision and demonstrate the denoising effect of anchor points in reducing these errors.
    \item We incorporate a smoothness penalty to enhance the temporal continuity of object trajectories.
    \item We validate the effectiveness of the proposed method through extensive experiments on simulated, real-world, and public datasets.
\end{itemize}

We organized the paper in the following manner. In Section \ref{sec:pre}, we introduce preliminary materials related to the camera imaging process and basic approaches for estimating target positions. In Section \ref{sec:anchor}, we demonstrate how the nominal localization method is affected by inaccurate camera parameters, propose the anchor-based localization method, and illustrate how this method can mitigate the influence of these inaccuracies. Additionally, we introduce a smoothness penalty to enhance temporal continuity. In Section \ref{sec:experiments}, we present extensive experiment results on simulated, real-world and public dataset. We believe that our approach is a showcase how optimization techniques can help to solve real-world engineering problems.

\section{Preliminaries} \label{sec:pre}

\subsection{Problem Setting}
Denote by $\mathbf{x}\in \mathbb{R}^{3}$ the 3D position of the target, and the position of $j^{th}$ anchor point in $k^{th}$ camera's view is $\mathbf{a}_{kj}\in \mathbb{R}^{3}$. Denote the pixel coordinate of target in $k^{th}$ camera's view by $\mathbf{y}_k\in \mathbb{R}^{3}$, and pixel coordinate of $j^{th}$ anchor point in $k^{th}$ camera's view by $\mathbf{b}_{kj}\in \mathbb{R}^{3}$. Let $\eta_k \in \{0,1\}$ denote whether the target is visible to $k^{th}$ camera. The parameters of camera $k$ includes, intrinsic matrix $\mathbf{M}_k\in \mathbb{R}^{3\times 3}$, rotation matrix $\mathbf{R}_k\in \mathbb{R}^{3\times 3}$, translation vector $\mathbf{T}_k\in \mathbb{R}^{3}$ and radial distortion coefficient $\mathbf{D}_k\in \mathbb{R}^{5}$. For simplicity, we denote by $\mathbf{h}_k$ the vectorized camera parameters, and the imaging process from 3D position to pixel coordinate is represented by the function $f(\mathbf{x},\mathbf{h})$, which is given by \eqref{eq:def:f} in Section \ref{sec: img2pix}, where we elaborate on the preliminaries of imaging process. To distinguish the ground-truth parameter and the calibrated ones, We use different upper-scripts for them. For example, $\widehat{\mathbf{h}}_k$ is the calibrated parameter and $\mathbf{h}_k^\star$ is the assumed ground-truth parameter. In total, there are $N_c$ cameras, and each camera is equipped with $N_a$ anchors.

Now we can model the observed pixel coordinate as
\begin{equation}
    \overline{\mathbf{y}}_k = f(\mathbf{x^\star},\mathbf{h}_k^\star) + \delta_k \label{eq:bar-y}
\end{equation}
where $\delta_k$ is pixel extraction error, for instance, error introduced by substituting head pixel with the upper central point of detection box. Similarly, the observed pixel coordinate of anchor point is 
\begin{equation}
    \overline{\mathbf{b}}_{kj} = f(\mathbf{a}_{kj}^\star,\mathbf{h}_k^\star) + \xi_{kj} \label{eq:bar-b}
\end{equation}
where $\xi_{kj}$ is the pixel extraction error of anchor points and is usually assumed to be much smaller than $\delta_k$. In addition, the reprojected anchor pixel using calibrated parameters is 
\begin{equation}
    \widehat{\mathbf{b}}_{kj} = f(\mathbf{a}_{kj}^\star,\widehat{\mathbf{h}}_k) \label{eq:hat-b}
\end{equation}

\subsection{Imaging Process}
\label{sec: img2pix}

In this section, we introduce preliminaries on imaging process considering radial distortion of lens. Given a 3D point in world coordinate system, \(\mathbf{x}=[x_1, x_2, x_3]^T\). \(\mathbf{x}\) can be transformed into camera coordinate system by applying extrinsics $\mathbf{R}$ and $\mathbf{T}$. After that, \([c_1, c_2, c_3]^T\) is homogenized by dividing $c_3$.
\begin{align}
    \begin{bmatrix} c_1 \\ c_2 \\ c_3 \end{bmatrix} = \mathbf{R} \times \begin{bmatrix} x_1 \\ x_2 \\ x_3 \end{bmatrix} + \mathbf{T},\  \ \begin{bmatrix} r_1 \\ r_2 \\ 1 \end{bmatrix}   =  \frac{1}{c_3}\begin{bmatrix} c_1 \\ c_2 \\ c_3 \end{bmatrix}.
\end{align}
After projected onto 2D image plane, distortion is applied to the projected image point \( \mathbf{r} = [r_1, r_2]\). Let $\mathbf{D}=[k_1, k_2, p_1, p_2 ,k_3]$ denote the radial distortion coefficients. The distortion process is denoted by $g: \mathbb{R}^7 \mapsto \mathbb{R}^2$ with $[r'_1, r_2'] = g(\mathbf{r}, \mathbf{D})$ given by
\begin{align}
\begin{split}
    r_1' & = r_1(1 + k_1 r_3^2 + k_2 r_3^4 + k_3 r_3^6) + (2p_1 r_1 r_2 + p_2(r_3^2 + 2 r_1^2))\\
    r_2' & = r_2(1 + k_1 r_3^2 + k_2 r_3^4 + k_3 r_3^6) + (p_1 (r_3^2 + 2 r_2^2) + 2 p_2 r_1 r_2)
    \end{split}
\end{align}
where $r_3^2 = r_1^2 + r_2^2$. Then,  
\([r_1', r_2', 1]^T\) in the camera coordinate system is transformed to pixel coordinate via the following process
\begin{align}
    \begin{bmatrix} u \\ v \\ 1 \end{bmatrix} = \mathbf{M} \times \begin{bmatrix} r_1' \\ r_2' \\ 1 \end{bmatrix},\ \ 
\mathbf{M} = \begin{bmatrix} f_x & 0 & c_x \\ 0 & f_y & c_y \\ 0 & 0 & 1 \end{bmatrix} \label{eq:pixel transform}
\end{align}
where $(f_x,f_y)$ and $(c_x,c_y)$ denote the focal length and principal point, respectively.  
The pixel coordinate is $\mathbf{y}=[u, v, 1]^T$. In summary, let $f$ denotes the imaging process from world coordinates $\mathbf{x}$ to pixel coordinate $\mathbf{y}$, then $f$ is given by:
\begin{align}\label{eq:def:f}
    \mathbf{y} = f(\mathbf{x},\mathbf{h}) = \mathbf{M} \cdot g\left( \frac{(\mathbf{R}\mathbf{x}+\mathbf{T})}{\mathbf{e_3^T}(\mathbf{R}\mathbf{x}+\mathbf{T})}, \mathbf{D} \right)
\end{align}

\subsection{Initial Estimation} \label{sec:init-est}
For initial estimation, we use homography mapping to project the pixels from different camera views onto a common ground plane or some fixed-height plane and then calculate the average position, which we refer to as \(\overline{\mathbf{x}}\). Specifically, let $\mathbf{H}_k$ be the homography matrix that transforms coordinates from the image plane of the $k^{th}$ camera to the top-down location on the common ground plane. Then for a given pixel $\mathbf{y}_k$, the corresponding grounding point is calculated by $\mathbf{H}_k \mathbf{y}_k$. When the target's ankle point is visible in the camera's FOV, this approach can directly applied on ankle pixel to determine target's location.

Under single-camera case, the initial point is chosen as the projected the ankle point on the ground plane. Suppose the target is only visible to the $k^{th}$ camera:
\begin{align}
    \overline{\mathbf{x}} = \mathbf{H}_k \mathbf{y}_k
\end{align}
For multi-camera case, we project the observed pixels from all camera views and project them onto the grounding plane, then the mean value is selected as initial estimation:
\begin{align}
    \overline{\mathbf{x}} = \frac{\sum_{k=1}^{N_c}\eta_k \mathbf{H}_k \mathbf{y}_k}{\sum_{k=1}^{N_c}\eta_k}
\end{align}
Note that in some tracking algorithms, this approach is used for integrating labeled targets into 3D locations \cite{luna2022online}.

\section{Anchor-based Localization Framework} \label{sec:anchor}

\subsection{Nominal Method} \label{sec:no-anchor-model}
In this section, we introduce the nominal localization method without anchor adjustments and analyze how it is influenced by camera parameter errors. The key idea is to determine a "hypothesized" 3D position $\mathbf{x}$ that ensures the reprojected pixels align as closely as possible with the observed pixels, or equivalently, minimizes the reprojection error. The corresponding optimization problem is formulated as a general least-square problem:

\begin{align}
 \min_{\mathbf{x}} &\quad  \sum _{k=1}^{N_c} \eta_k \|\widehat{\mathbf{y}}_k(\mathbf{x}) - \overline{\mathbf{y}}_k\|^2_2 \label{prog: basic_model}\\
\text{where}  & \quad \widehat{\mathbf{y}}_k(\mathbf{x})  \triangleq f(\mathbf{x}, \widehat{\mathbf{h}}_k) 
\label{eq:y_hat} 
\end{align}
where $\overline{\mathbf{y}}_k$ is the observed pixel coordinate defined in \eqref{eq:bar-y}. When $N_c=1$, we fix the height as a constant and optimize only the $x$ and $y$ coordinates. 

Ideally, minimizing the objective function should yield a solution close to the ground-truth location $\mathbf{x^\star}$. However, due to calibration and pixel extraction errors, the solution to \eqref{prog: basic_model} may deviate from the true location $\mathbf{x^\star}$. Using Taylor expansion, we have
\begin{align} \label{target expansion}
    \begin{split}
\widehat{\mathbf{y}}_k(\mathbf{x}) - \overline{\mathbf{y}}_k = & f(\mathbf{x}, \widehat{\mathbf{h}}_k)-f(\mathbf{x}^\star, \mathbf{h}_k^\star) - \delta_k \\ 
    =& f(\mathbf{x}, \mathbf{h}_k^\star) - f(\mathbf{x}^\star, \mathbf{h}_k^\star) + f(\mathbf{x}, \widehat{\mathbf{h}}_k)- f(\mathbf{x}, \mathbf{h}_k^\star) - \delta_k \\
=&\nabla_{x}f(\mathbf{x}^\star,\mathbf{h}_k^\star)^T(\mathbf{x}-\mathbf{x}^\star) + \nabla_{\mathbf{h}}f(\mathbf{x},\mathbf{h}_k^\star)^T(\widehat{\mathbf{h}}_k-\mathbf{h}_k^\star) - \delta_k  \\ 
    & + o(\|\widehat{\mathbf{h}}_k-\mathbf{h}_k^\star\| + \|\mathbf{x}-\mathbf{x}^\star\|)  \end{split}
\end{align}
Denote by $\epsilon_h^k$ the error term $\nabla_{\mathbf{h}}f(\mathbf{x},\mathbf{h}_k^\star)^T(\widehat{\mathbf{h}}_k-\mathbf{h}_k^\star)$, arising from inaccuracy of $k$-th camera parameters. By ignoring the higher-order term in the Little-o notation, we express the objective as:
$$
\sum _{k=1}^{N_c} \eta_k \|\widehat{\mathbf{y}}_k(\mathbf{x}) - \overline{\mathbf{y}}_k\|^2_2 = \left\| \mathbf{H}(\mathbf{x}- \mathbf{x}^\star) + \mathbf{\epsilon}_h - \delta\right\|_2^2
$$
where the $k$-th column of $\mathbf{H}$ corresponds to $\nabla_{x}f(\mathbf{x}^\star, \mathbf{h}_k^\star)$. Assuming $\mathbf{H}$ has full row rank, the optimal solution to \eqref{prog: basic_model} is given by 
\begin{align}\label{eq:nominal-solution}
\mathbf{x} = \mathbf{x}^\star - (\mathbf{H}\mathbf{H}^T)^{-1} \mathbf{H} (\epsilon_h -\delta).
\end{align}
As shown, the estimation error consists of two components: (1) inaccuracies in the camera parameters $\epsilon_h$ and (2) pixel extraction errors $\delta$. In practice, errors from inaccurate camera parameters often dominate, causing the solution to \eqref{prog: basic_model} to deviate significantly from the ground-truth position $\mathbf{x}^\star$.

\subsection{Anchor-based Method} \label{sec:anchor-model}
In this section, we propose an anchor-based localization method aimed at minimizing anchor-adjusted reprojection errors:
\begin{align}\label{prog: anchor_model}
    \min_{\mathbf{x}} & \quad\sum_{k=1}^{N_c}  \eta_k \left\|  \widehat{\mathbf{y}}_k(\mathbf{x}) - \overline{\mathbf{y}}_k - \sum_{j=1}^{N_a} \omega_{kj}\left(\widehat{\mathbf{b}}_{kj} - \overline{\mathbf{b}}_{kj}\right) \right\|^2_2
\end{align}
where $\widehat{\mathbf{y}}_k(\mathbf{x}), \overline{\mathbf{y}}_k, \widehat{\mathbf{b}}_{kj}$, and $ \overline{\mathbf{b}}_{kj}$ are defined in \eqref{eq:y_hat}, \eqref{eq:bar-y}, \eqref{eq:hat-b}, and \eqref{eq:bar-b}, respectively. The weights $\{\omega_{kj}\}$ must be carefully selected. In the following, we provide a theoretical framework for determining these weights to effectively reduce calibration errors by leveraging anchor points.

To begin, consider the reprojection error of the $j^{th}$ anchor in the $k^{th}$ camera view:
\begin{align}
\begin{split}
    \widehat{\mathbf{b}}_{kj} - \overline{\mathbf{b}}_{kj} = & f(\mathbf{a}_{kj}^\star, \widehat{\mathbf{h}}_k)-f(\mathbf{a}_{kj}^\star, \mathbf{h}_k^\star) - \xi_{kj}\\
=&\nabla_{\mathbf{h}}f(\mathbf{a}_{kj}^\star, \mathbf{h}_k^\star) ^T(\widehat{\mathbf{h}}_k-\mathbf{h}_k^\star) + o(\|\widehat{\mathbf{h}}_k-\mathbf{h}_k^\star\|) - \xi_{kj} \label{anchor expansion}
\end{split}
\end{align}
where the second equality follows from the Taylor expansion. 
By ignoring the higher-order term in the Little-o notation and assuming pixel extraction errors on anchors are negligible, we observe that the first term in \eqref{anchor expansion} shares a similar structure with the corresponding term in \eqref{target expansion}, but with different coefficients. This similarity motivates us to further expand the coefficient term $\nabla_{\mathbf{h}}f(\mathbf{a}_{kj}^\star, \mathbf{h}_k^\star)$ at the point $(\mathbf{x}, \mathbf{h}_k^\star)$:
\begin{align}
\nabla_{\mathbf{h}}f(\mathbf{a}_{kj}^\star, \mathbf{h}_k^\star) = \nabla_{\mathbf{h}}f(\mathbf{x}, \mathbf{h}_k^\star) +\nabla_{x}\nabla_{\mathbf{h}}f(\mathbf{x},\mathbf{h}_k^\star )(\mathbf{a}_{kj}^\star-\mathbf{x}). \label{twice-dif}
\end{align}
Note that the first term on the right-hand side of \eqref{twice-dif} matches the first term in \eqref{target expansion}. To eliminate the second term in \eqref{twice-dif}, we can construct an affine combination of the anchor points with weights $\{\omega_{kj}\}_{j=1}^{N_a}$, such that 
\begin{align}\label{eq:weights}
    \sum_{j=1}^{N_a} \omega_{kj}=1, \quad \mathbf{x} = \sum_{j=1}^{N_a} \omega_{kj} \mathbf{a}_{kj}^\star.
\end{align}
Although the ground-truth position $\mathbf{x}$ is unknown, we can approximate the weights accordingly (see Section \ref{sec: estimate omega}).  Applying this affine combination allows us to simplify the expression, yielding
\begin{align}
    \sum_{j=1}^{N_a} \omega_{kj} \nabla_{\mathbf{h}}f(\mathbf{a}_{kj}^\star, \mathbf{h}_k^\star) = & \sum_{j=1}^{N_a} \omega_{kj} \nabla_{\mathbf{h}}f(\mathbf{x}, \mathbf{h}_k^\star) + \nabla_{x}\nabla_{\mathbf{h}}f(\mathbf{x},\mathbf{h}_k^\star ) \left(\sum_{j=1}^{N_a} \omega_{kj} \mathbf{a}_{kj}^\star-\mathbf{x}\right)\notag \\
    = & \nabla_{\mathbf{h}}f(\mathbf{x}, \mathbf{h}_k^\star). \label{weighted-anchor}
\end{align}
This suggests that the affine combination of anchor reprojection errors can effectively reduce the camera parameter error term $\nabla_{\mathbf{h}}f(\mathbf{x}, \mathbf{h}_k^\star)(\widehat{\mathbf{h}}_k-\mathbf{h}_k^\star)$. Next, combine \eqref{target expansion}, \eqref{anchor expansion}, \eqref{twice-dif} and \eqref{weighted-anchor} to obtain
\begin{align}
\begin{split}
    & \widehat{\mathbf{y}}_k(\mathbf{x}) - \overline{\mathbf{y}}_k - \sum_{j=1}^{N_a} \omega_{kj} \left(\widehat{\mathbf{b}}_{kj} - \overline{\mathbf{b}}_{kj}\right)  \\
    = & \nabla_{x}f(\mathbf{x}^\star, \mathbf{h}_k^\star)^T(\mathbf{x}-\mathbf{x}^\star) -\delta_k+\sum_{j=1}^{N_a}\omega_{kj}\xi_{kj}+ o(\|\widehat{\mathbf{h}}_k-\mathbf{h}_k^\star\| + \|\mathbf{x}-\mathbf{x}^\star\|). \label{eq: anchor_error_term}
    \end{split}
\end{align}
Finally, summing these terms and ignoring the higher-order term in the Little-o notation yield:
\begin{align*}
    \begin{split}
        \sum_{k=1}^{N_c}  \left\|  \widehat{\mathbf{y}}_k(\mathbf{x}) - \overline{\mathbf{y}}_k - \sum_{j=1}^{N_a} \omega_{kj}\left(\widehat{\mathbf{b}}_{kj} - \overline{\mathbf{b}}_{kj}\right) \right\|^2_2 = \left\| \mathbf{H}^T(\mathbf{x} - \mathbf{x}^\star) - \delta + \xi \right\|^2_2 
    \end{split}
\end{align*}
where the $k$-th entry of $\xi$ is given by $\sum_{j=1}^{N_a}\omega_{kj}\xi_{kj}$. Assuming $\mathbf{H}$ has full row rank, the optimal solution to \eqref{prog: anchor_model} is given by
\begin{align}\label{eq:anchor-solution}
    \mathbf{x} = \mathbf{x}^\star - (\mathbf{H}\mathbf{H}^T)^{-1} \mathbf{H} (\xi - \delta).
\end{align}
Compared with \eqref{eq:nominal-solution}, \eqref{eq:anchor-solution} successfully eliminates the error arising from inaccuracies in the camera parameters.  Note that the anchor points also introduce extra pixel extraction error term $\sum_{j=1}^{N_a}\omega_{kj}\xi_{kj}$. In practice, this error term $\xi_{kj}$ is typically much smaller than $\delta_k$. This derivation rigorously demonstrates how the formulation in \eqref{prog: anchor_model} reduces the systematic calibration error term in \eqref{target expansion}.

\subsubsection{Estimation of the weights $\omega$}\label{sec: estimate omega}
For anchor-based formulation \eqref{prog: anchor_model}, the weights $\{\omega_{kj}\}$ need to satisfy the constraints in \eqref{eq:weights}. Although the ground-truth position $\mathbf{x}$ is unknown, we can approximate the weights using initial estimation point \(\overline{\mathbf{x}}\). Specifically, the weights should satisfy:
\begin{align}
\sum_{j=1}^{N_a} \overline{\omega}_{kj}=1,\quad  \overline{\mathbf{x}} = \sum_{j=1}^{N_a} \overline{\omega}_{kj}\mathbf{a}_{kj}^\star. \label{constr:approx}
\end{align}
Notice that without additional constraints, solving \eqref{constr:approx}  may yield solutions with large absolute values for some entries, potentially leading to numerical instability. 

To address this issue, we modify the problem by introducing a penalty term for the weights  $\overline{\omega}$. The reformulated problem is:
\begin{align}
\begin{split}
\min_{\overline{\omega}_{kj}} & \quad  \left\| \overline{\mathbf{x}} - \sum_{j=1}^{N_a} 
\overline{\omega}_{kj}\mathbf{a}_{kj}^\star\right\|_2^2 + \lambda \|\overline{\omega}_k\|^2_2 \label{prog:w-penalty}\\
\text{s.t.} & \quad \sum_{j=1}^{N_a} \overline{\omega}_{kj}=1.
\end{split}
\end{align}

\subsection{Localization with Smoothness Penalty} \label{sec: smooth}
Given that the localization target forms a trajectory, it is essential to leverage temporal information across frames to enhance localization precision. Temporal continuity plays a crucial role, especially in real-world scenarios where frequent occlusions occur. As objects move, the number of cameras observing the target often changes, causing localization precision to drop—particularly when the target transitions from a cross-view region (covered by multiple cameras) to a region visible by only a single camera. This leads to non-smooth trajectories. 

To address this, we propose a localization framework with a smoothness penalty. The framework determines the target's locations within a specified time window, simultaneously minimizing both the anchor-adjusted reprojection error and a smoothness penalty for changes in consecutive frames. Assuming the time horizon (or batch size) spans $T$ frames, the optimization problem is formulated as:
\begin{align}
    \min_{\mathbf{x_1,..,x_T}} & \sum_{t=1}^{T} \sum_{k=1}^{N_c}  \left\|  \widehat{\mathbf{y}}_k^t(\mathbf{x}_t) - \overline{\mathbf{y}}_k^t - \sum_{j=1}^{N_a} \overline{\omega}^t_{kj}\left(\widehat{\mathbf{b}}_{kj} - \overline{\mathbf{b}}_{kj}\right) \right\|^2_2 + \sum_{t=2}^{T} \rho \|\mathbf{x}_t-\mathbf{x}_{t-1}\|^2_2\label{eq:rho}
\end{align}
where $\overline{\mathbf{y}}_k^t, \widehat{\mathbf{y}}_k^t(\mathbf{x}_t)$ represent the hypothesized and observed pixel of target under camera $k$ in frame $t$, and $\overline{\omega}^t_{kj}$ satisfies $\sum_{j=1}^{N_a} \overline{\omega}^t_{kj}=1, \overline{\mathbf{x}}_t = \sum_{j=1}^{N_a} \overline{\omega}^t_{kj}\mathbf{a}_{kj}^\star$. 

\section{Experiments} \label{sec:experiments}
To validate the effectiveness of the proposed anchor-based localization method, we conduct extensive numerical experiments on simulated (Section \ref{sec: exp}), real-world (Section \ref{sec:exp-real}), and public dataset (Section \ref{sec:exp-public}). Our code is available at \href{https://github.com/zwyhahaha/AnchorLocalization.git}{AnchorLocalization}.

\subsection{Experiments on Simulated Data} \label{sec: exp}

\subsubsection{Simulation Setting}

\begin{figure}
  \centering
  \begin{tabular}{cc}
    \includegraphics[width=0.45\linewidth]{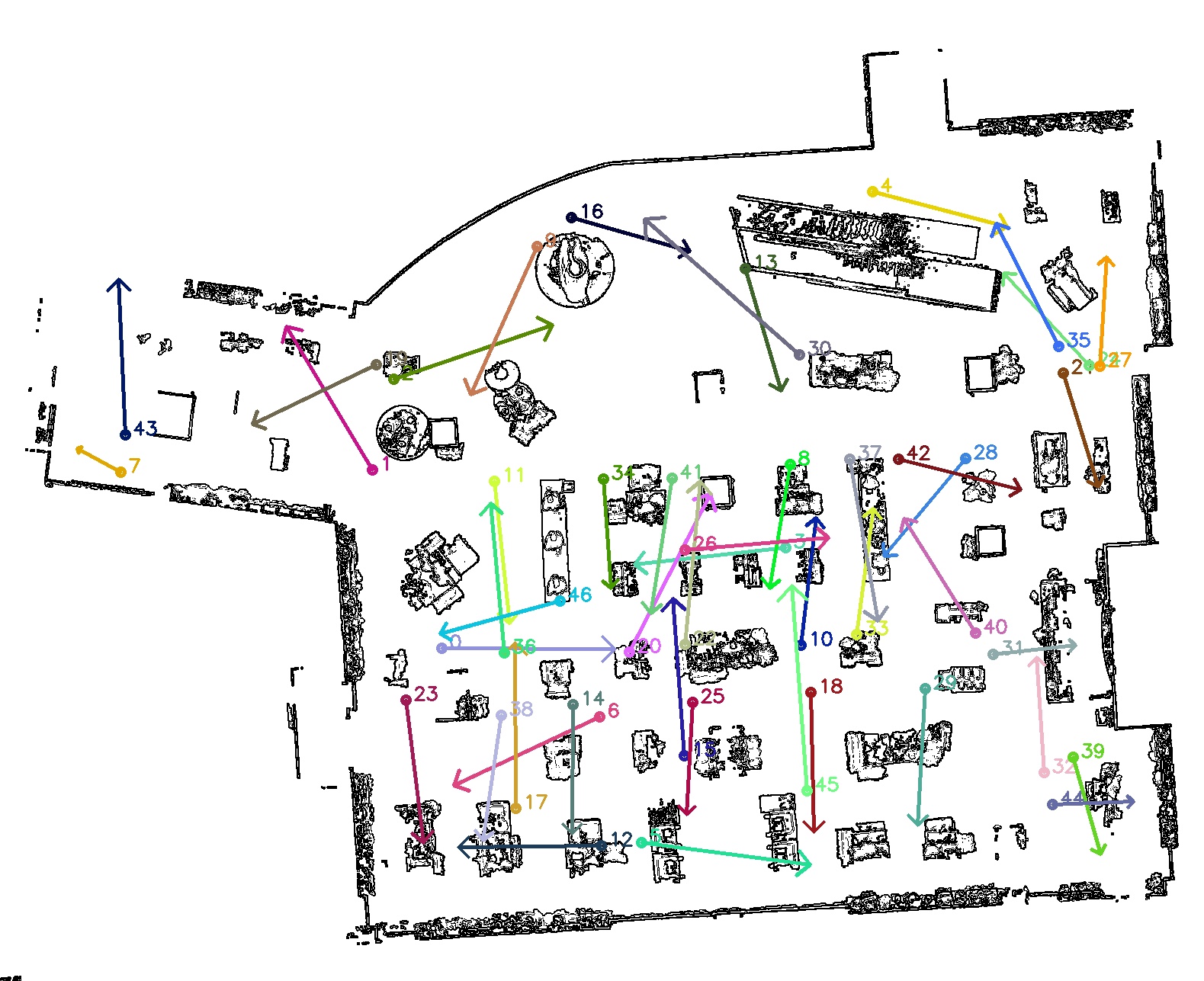} &
    \includegraphics[width=0.45\linewidth]{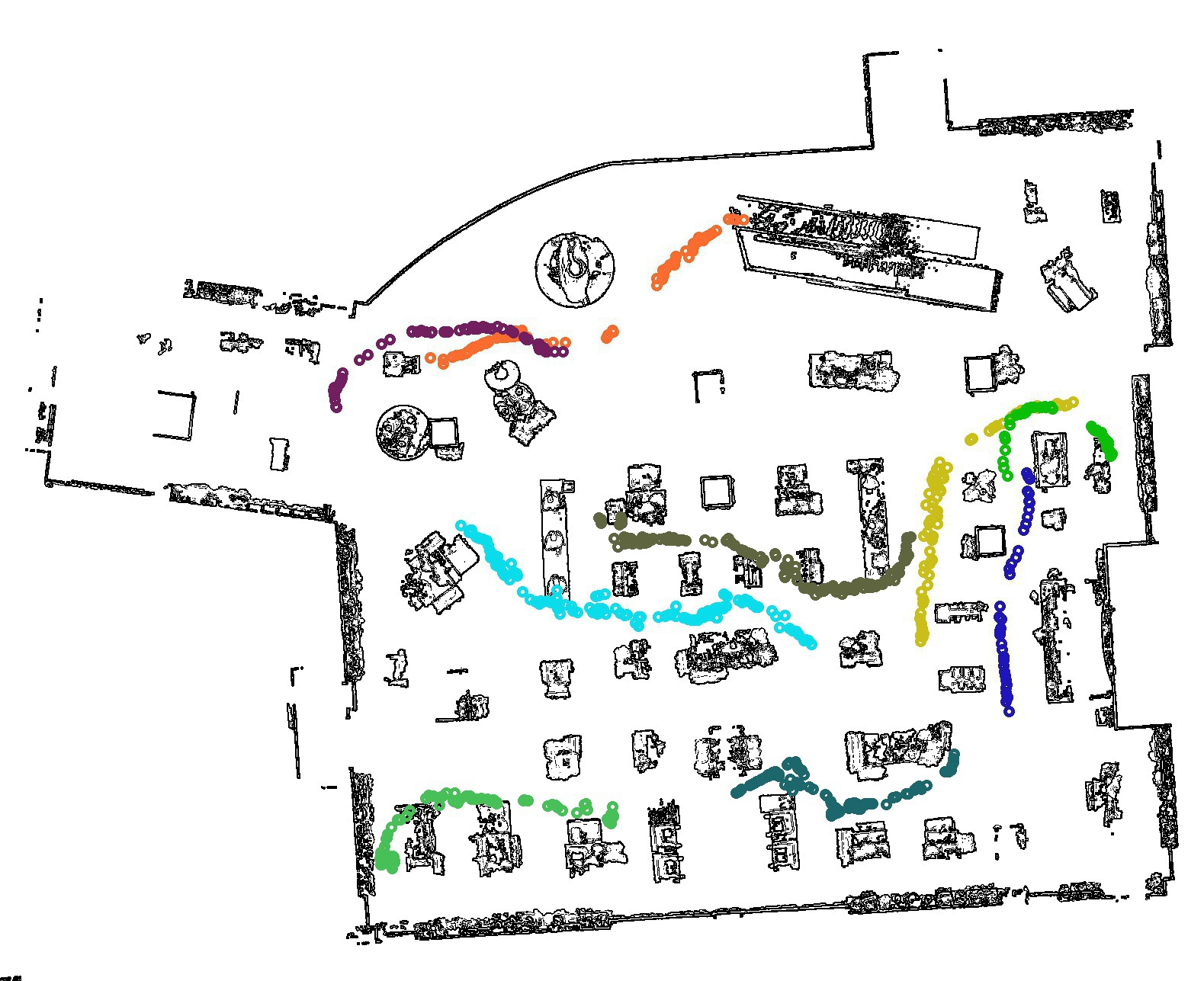}
  \end{tabular}
  \caption{\emph{Left:} The 2D map of simulation environment. The circles represent the locations of each camera while the corresponding lines are their orientations. \emph{Right: } The simulated trajectories.}
  \label{fig:simu-environment}
\end{figure}

The simulation environment is based on a real-world scenario provided by \href{https://cue.group/en/}{CUE Group}. This scenario features a rectangular space measuring 49 meters in length and 39 meters in width, covered by 47 calibrated cameras. The calibration is conducted by a standard method provided by  \href{https://docs.opencv.org/4.x/dc/dbb/tutorial_py_calibration.html}{opencv} library. The map of the scenario, along with the positions and orientations of the cameras, is illustrated in the graph to the left of Figure \ref{fig:simu-environment}.

% \vspace{-5mm}
% \vspace{-5mm}

\paragraph{Trajectory Simulation.} We generate a set of ground truth trajectories with 1000 frames based on the random-walk model and generate the height of each trajectory randomly. Also, we perceive given camera parameters as the ground-truth. Subsequently, we calculate the observed pixels by employing the imaging process outlined in Appendix \ref{sec: img2pix}, utilizing ground-truth camera parameters. Random noise is then introduced to the calculated pixels to simulate pixel extraction errors. Regarding anchor points, we uniformly sample the positions of 10 anchors within the field of view (FOV) of each camera, and their ground-truth world coordinates are available for the localization procedure. The simulated trajectories are shown in the right side of Figure \ref{fig:simu-environment}.

\paragraph{Camera Parameter Simulation.} To simulate camera parameter errors, we introduce perturbations to given camera parameters. For rotation matrix, we first calculate the camera's pitch, yaw, and roll angles and apply additive noise on the pitch and yaw angles (termed Rx and Ry errors), since a slight change in the roll angle would introduce abnormal tilt to the camera's image thus is easy to detect in practice. Translation vector, which describes the camera's shift from the the world coordinate system origin, is applied with additive noise (T error). Lastly, we apply multiplicative noise on distortion coefficients (D error). The simulation error setting and corresponding angular errors (in degree) are listed in Table \ref{tab: pix and ang error}.

\begin{table}[]
\centering
\caption{Camera parameter errors and corresponding angular errors (in degree)}
\begin{tabular}{c@{\hspace{0.3cm}}c@{\hspace{0.4cm}}c|c@{\hspace{0.3cm}}c@{\hspace{0.4cm}}c}
\toprule
\multicolumn{1}{c}{Term} & \multicolumn{1}{c@{\hspace{0.35cm}}}{Value} & \multicolumn{1}{c|}{Angular error} & \multicolumn{1}{c}{Term} & \multicolumn{1}{c@{\hspace{0.35cm}}}{Value} & \multicolumn{1}{l}{Angular error} \\ 
\midrule
% No error &0.0 &0.066    & & &\\
% \hline
\multirow{6}{*}{\begin{tabular}[c]{@{\hspace{0.4cm}}c@{}}Rx error \\ (degree)\end{tabular}}     
& 0.25  & 0.350           & \multirow{6}{*}{\begin{tabular}[c]{@{\hspace{0.4cm}}c@{}}Ry error \\ (degree)\end{tabular}}      & 0.25  &0.530           \\
& 0.50  & 0.650           & & 0.50  &1.000           \\
& 0.75  & 0.950           & & 0.75  &1.470           \\
& 1.00  & 1.250           & & 1.00  &1.940           \\
& 1.25  & 1.570           & & 1.25  &2.398           \\
& 1.50  & 1.890           & & 1.50  &2.858           \\
\hline
\multirow{5}{*}{\begin{tabular}[c]{@{\hspace{0.4cm}}c@{}}T error \\ (meter)\end{tabular}}      & 0.05  & 0.380           & \multirow{5}{*}{\begin{tabular}[c]{@{\hspace{0.4cm}}c@{}}D error
       \\ 
\end{tabular}}      & 0.05  &0.164           \\
& 0.10  & 0.675           & & 0.10  &0.273           \\
& 0.15  & 1.041           & & 0.15  &0.386           \\
& 0.20  & 1.422           & & 0.20  &0.501           \\
& 0.25  & 1.609           & & 0.25  &0.623           \\
\bottomrule
\end{tabular}
\label{tab: pix and ang error}
\end{table}

% \vspace{-0.5mm}

\paragraph{Evaluation Metrics.} To benchmark different localization methods, we use the average distance (in meters) between the localization results and the ground-truth positions as a performance measure. Additionally, we employ the improvement ratio, defined as the percentage of instances in which localization precision improves after optimization compared to the initial estimation, to assess the method's stability.

\subsubsection{Robustness to Camera Parameter Errors}
\begin{figure}[htb]
    \centering
    \includegraphics[width=\linewidth]{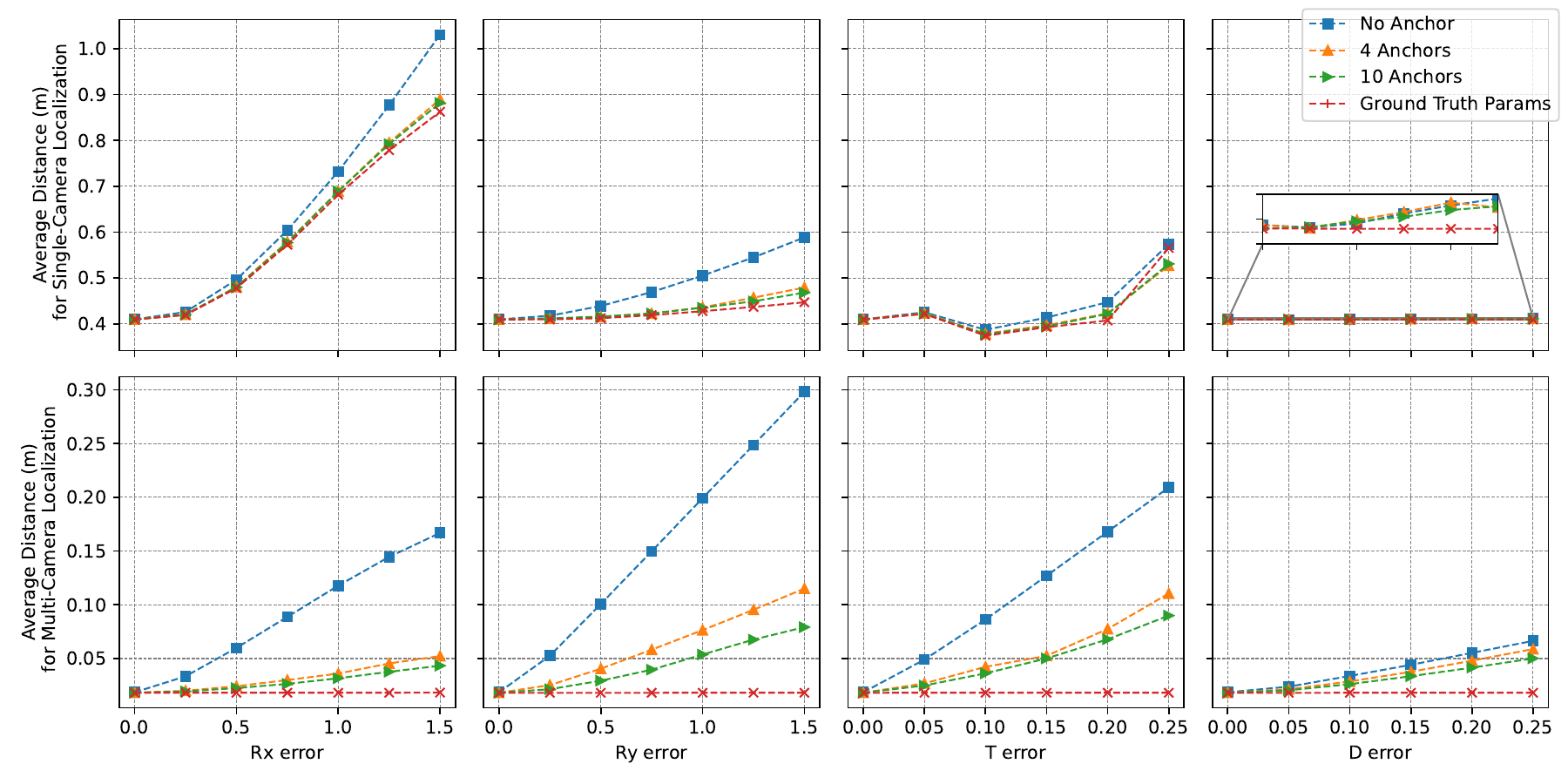}
    \caption{Average Distance (m) of different localization methods under camera parameter errors in Table \ref{tab: pix and ang error}. The data point is the average of both positive and negative perturbations.}
    \label{fig:simu-cam-err-loc}
\end{figure}

\begin{figure}[htb]
    \centering
    \includegraphics[width=\linewidth]{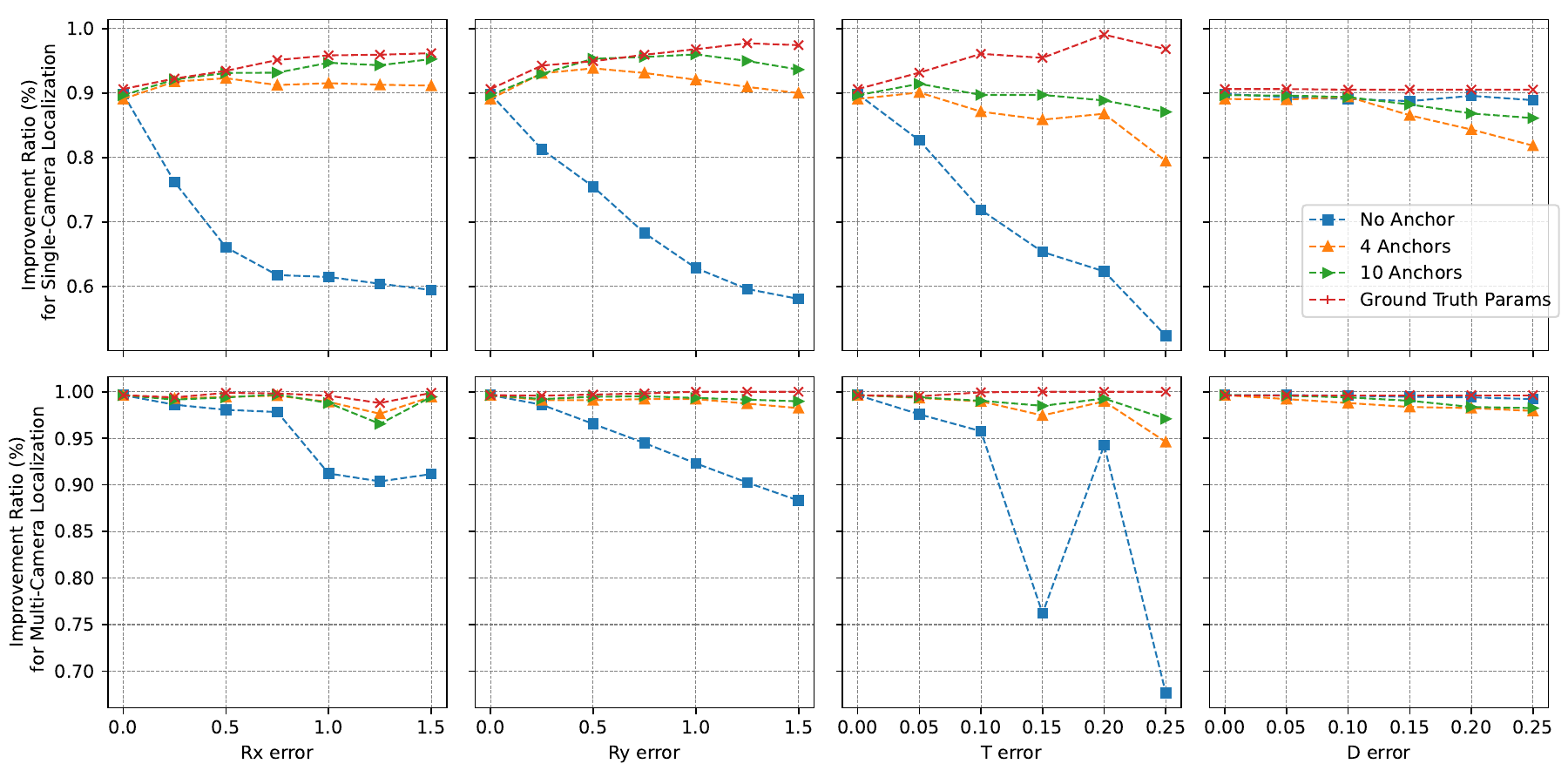}
    \caption{Improvement Ratio of different localization methods under camera parameter errors in Table \ref{tab: pix and ang error}. The data point is the average of both positive and negative perturbations.}
    \label{fig:simu-cam-err-loc-imp}
\end{figure}

% To justify the robustness of proposed method to camera calibration errors, 
We compare the proposed anchor localization method with the no anchor method under different parameter error settings, as shown in Table \ref{tab: pix and ang error}. For anchor-based method, we vary anchor numbers and set $N_a=4,10$. To facilitate better benchmarking, we test no anchor method with the ground-truth camera parameters, representing the best achievable results. The initial solution is the same for all methods, and the pixel perturbation level is set as 3 pixels. The experimental results, evaluated based on average distance and improvement ratio, are presented in Figure \ref{fig:simu-cam-err-loc} and Figure \ref{fig:simu-cam-err-loc-imp}. The experiment results for combinatorial camera parameter errors are deferred to Table \ref{tab:combination of errors} in Appendix \ref{app:exp-simu}.

% \vspace{-0.5mm}

From the experimental results, we observe that, across various camera parameter errors, the anchor-based method achieves a lower average localization distance and a higher improvement ratio compared to the nominal model. As the calibration error increases, the performance of all localization methods degrades (in ground-truth experiments, the localization precision is affected by inaccurate initial estimations). In all scenarios, the anchor-based method consistently achieves stable performance, with an improvement ratio greater than 90\%, demonstrating robustness to camera parameter errors. Furthermore, we note that even in a single-camera localization setting, the anchor-based method can still reduce the localization distance. An increasing number of anchors further nudges the localization precision closer to the best possible results obtained using ground-truth parameters.

\subsection{Experiments on Real-World Data} \label{sec:exp-real}

\subsubsection{Real-World Data Setting} 
\begin{figure}
  \centering
  \begin{tabular}{ccc}
    \includegraphics[width=0.3\linewidth]{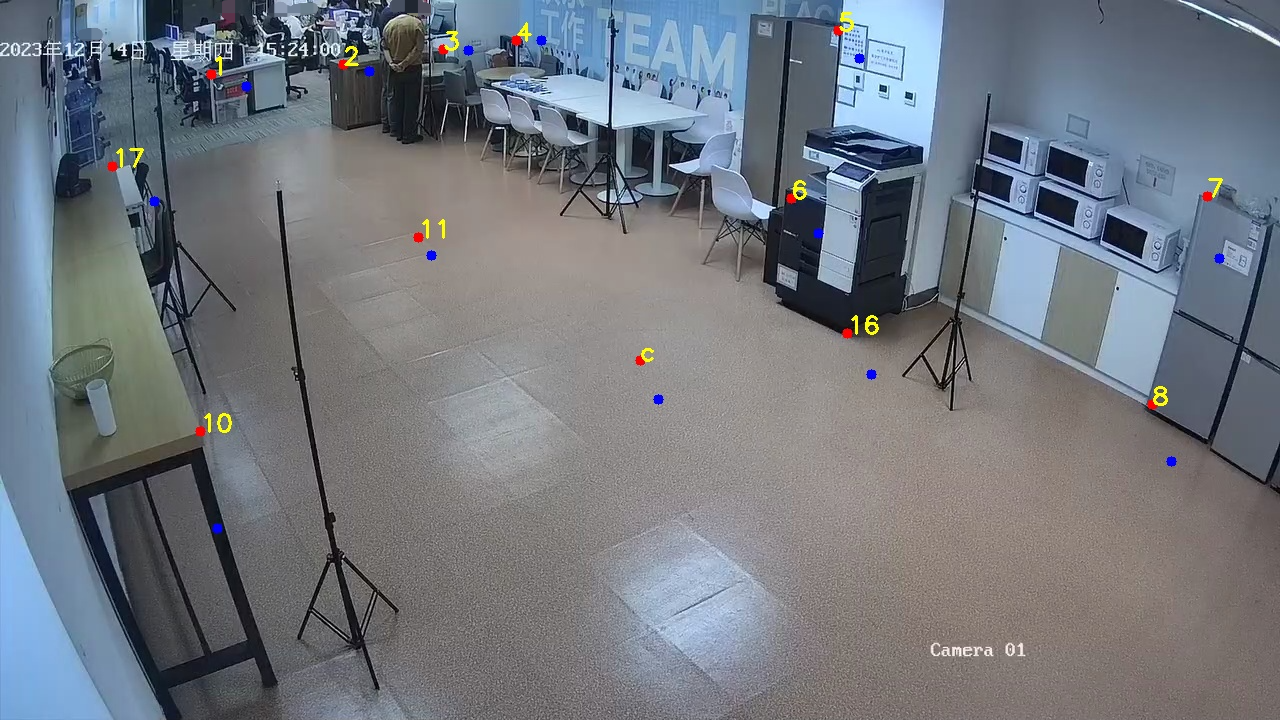} &
    \includegraphics[width=0.3\linewidth]{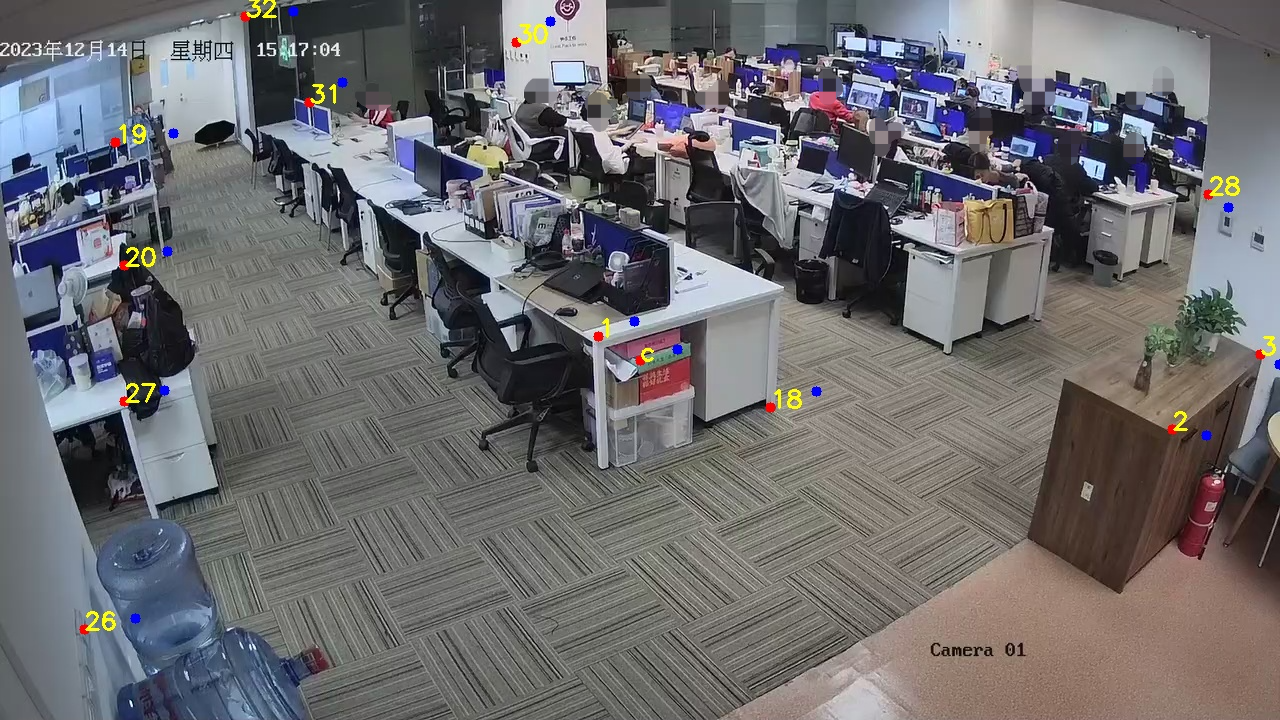} &
    \includegraphics[width=0.3\linewidth]{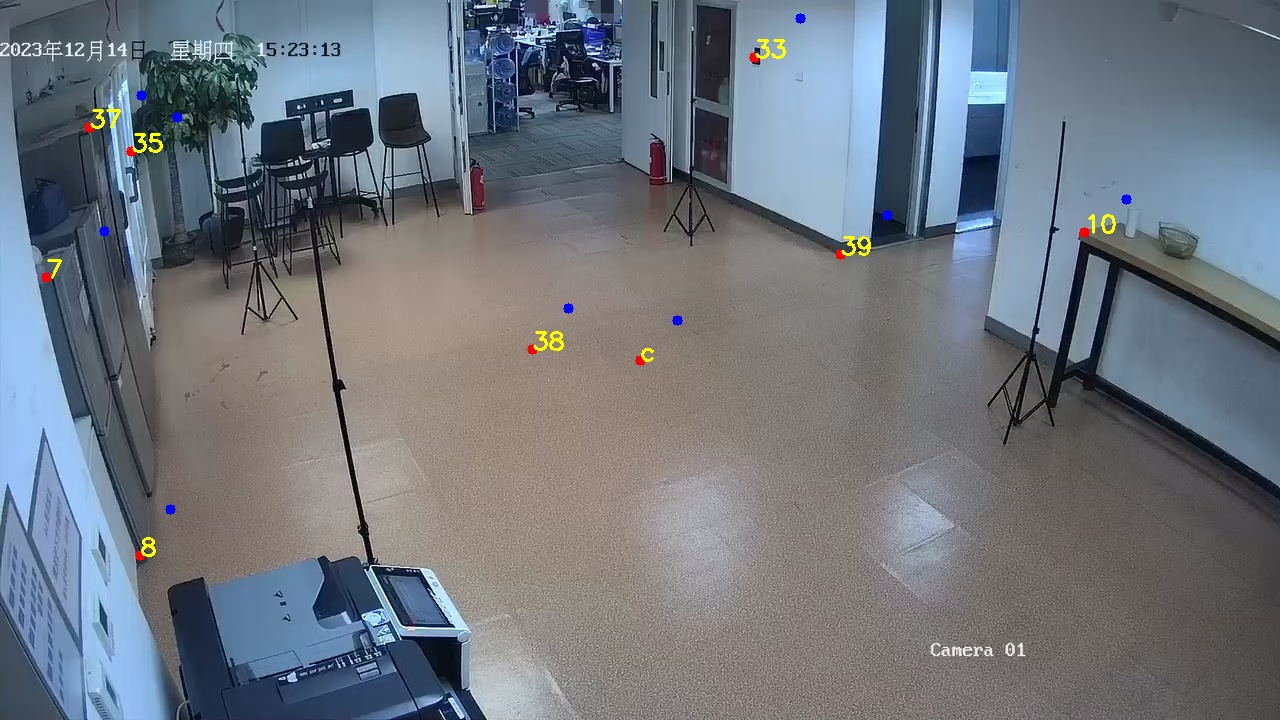} \\
    \includegraphics[width=0.3\linewidth]{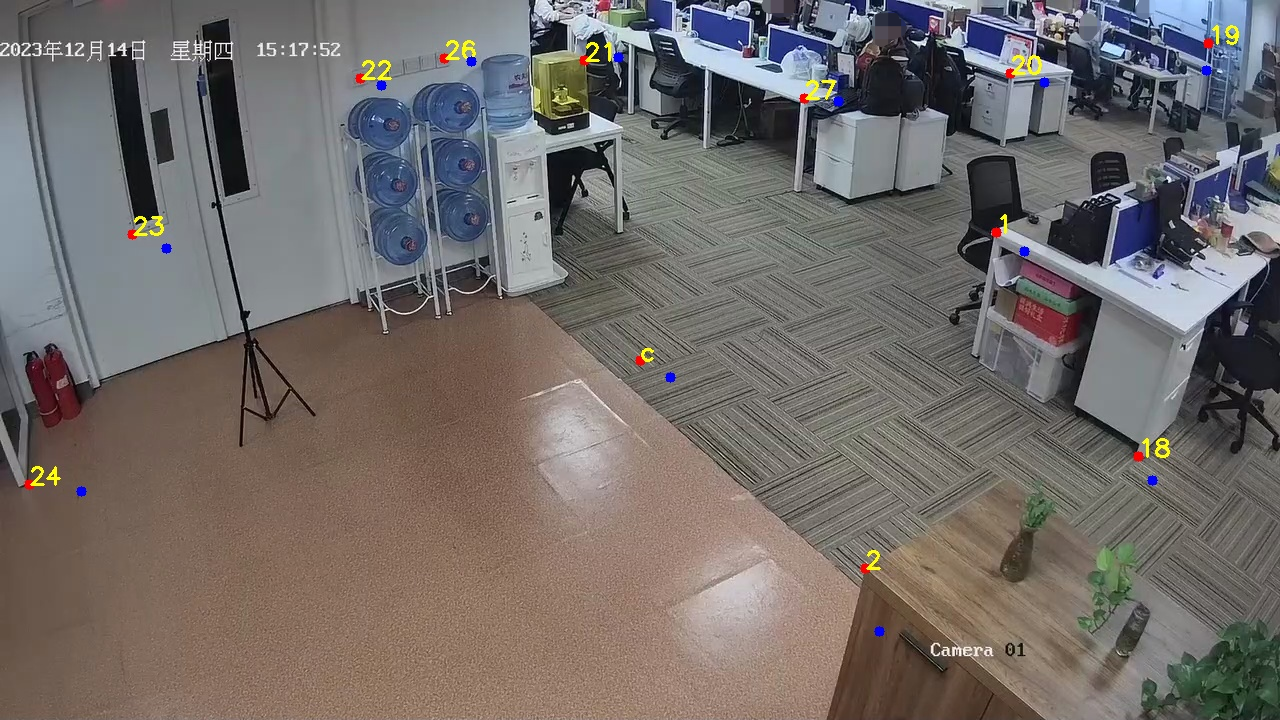} &
    \includegraphics[width=0.3\linewidth]{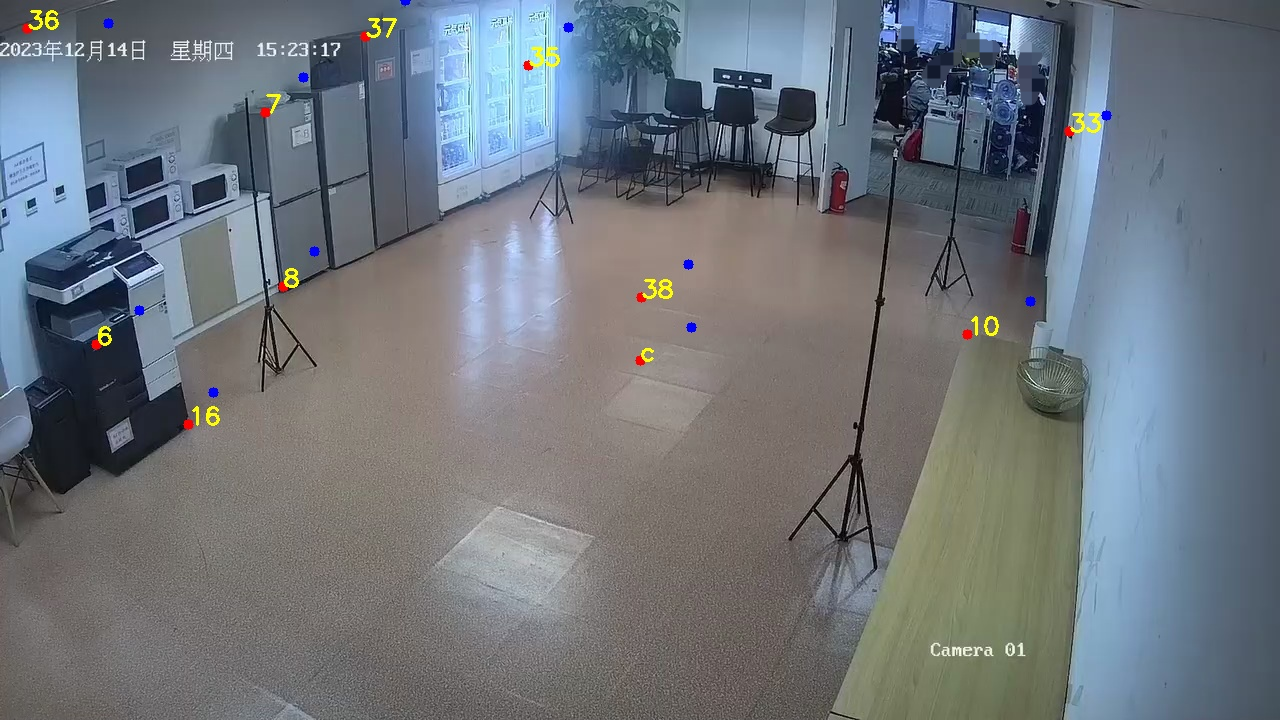} &
    \includegraphics[width=0.3\linewidth]{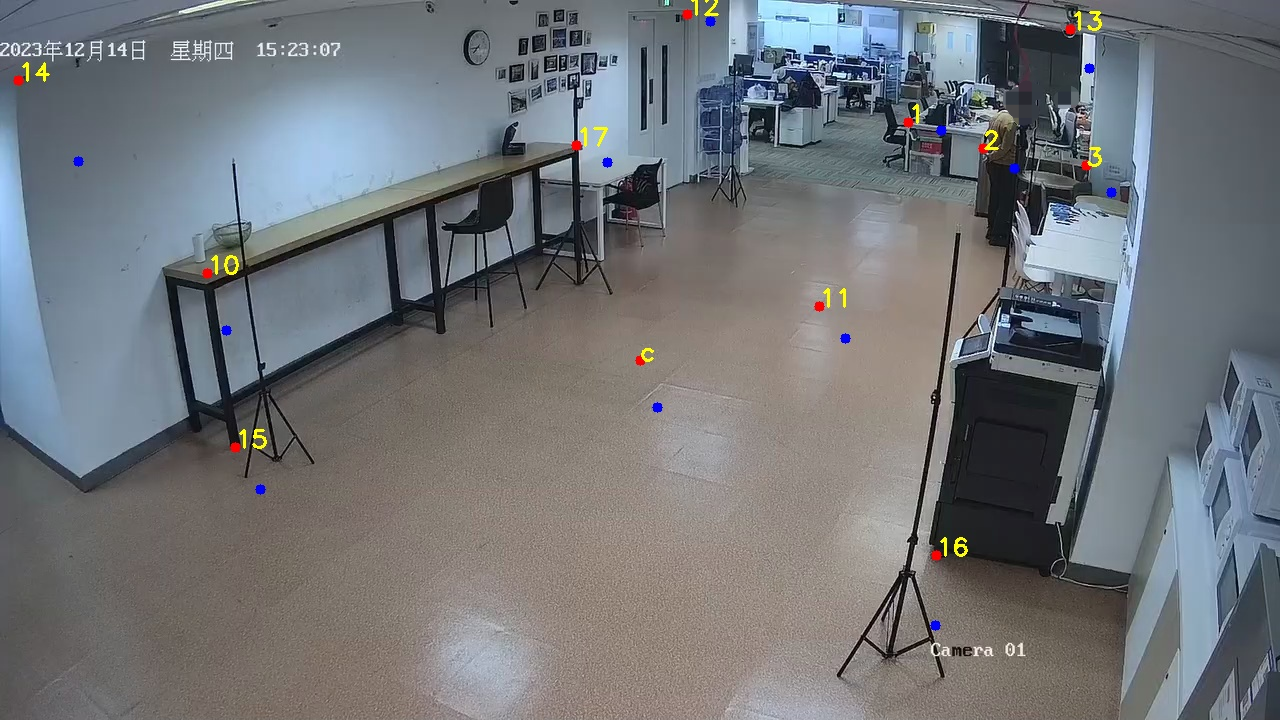}
  \end{tabular}
  
  \caption{Selected anchors points of 6 cameras. Red points represent the anchor points observed in the images. Blue points represent the reprojected pixels from the surveyed anchor positions.}
  \label{fig:anchor-example}
\end{figure}

The real-world data is provided by \href{https://cue.group/en/}{CUE Group}, a Chinese cooperation that provides technological solutions to industries such as smart commerce. The dataset includes the trajectories of 7 targets observed by 6 cameras. In addition to 1700 frames of live stream videos with a resolution of $1280 \times 720$, the dataset also provides measured ground-truth locations of these trajectories to benchmark the localization results. To ensure high precision of the ground-truth locations, UWB positioning is employed using 8 indoor sensors. All cameras are calibrated by a standard approach provided by  \href{https://docs.opencv.org/4.x/dc/dbb/tutorial_py_calibration.html}{opencv} library. The ground-truth positions of anchor points in its FOV are carefully measured. These anchor points are selected as the corners of conspicuous objects to accurately determine their 3D positions and observed pixels. Each camera is equipped with 8-12 anchors (refer to Figure \ref{fig:anchor-example} for examples). Detection boxes are labeled by \verb|YOLOv7| \cite{wang2023yolov7} with human correction.

\subsubsection{Ideal Anchor Localization Experiments} 
Since the ground-truth positions of the anchor points are known and these points are subject to small pixel extraction errors, they serve as ideal localization targets. The experiment results are presented in Table \ref{tab:anchor loc}. There are 16 anchors that are only visible to one camera, and their height is set to be known. In this scenario, the precision of the initial estimation is low because most of these anchors are located in severely distorted peripheral areas of the cameras. Nonetheless, the anchor-based method can improve precision to some extent. The remaining 19 anchors are visible to multiple cameras. In this case, the anchor-based method demonstrates high accuracy, with an average distance below 30 cm, which is approximately 20 cm lower than that of the nominal method. This improved accuracy can be attributed to the small pixel extraction errors associated with the anchors.
% \vspace{-0.5mm}
\begin{table}[ht]
    \centering
    \caption{Average distance (m) of different methods in anchor localization experiment.}
    \begin{tabular}{ccccc}
    \toprule
         &  Anchor Number &  Initial Estimation & No Anchor & Anchor-based \\
    \midrule
         Single Camera &  16 &  2.695 & 2.379 & 2.289 \\
         Multi Camera &  19 &  1.328 & 0.426 & 0.245 \\
    \bottomrule
\end{tabular}
    \label{tab:anchor loc}
\end{table}
% \vspace{-0.5mm}

\subsubsection{Pedestrian Localization Experiment} \label{sec:exp-pedestrian}

\begin{figure}[]
  \centering
  \begin{subfigure}{0.45\linewidth}
    \includegraphics[width=\linewidth]{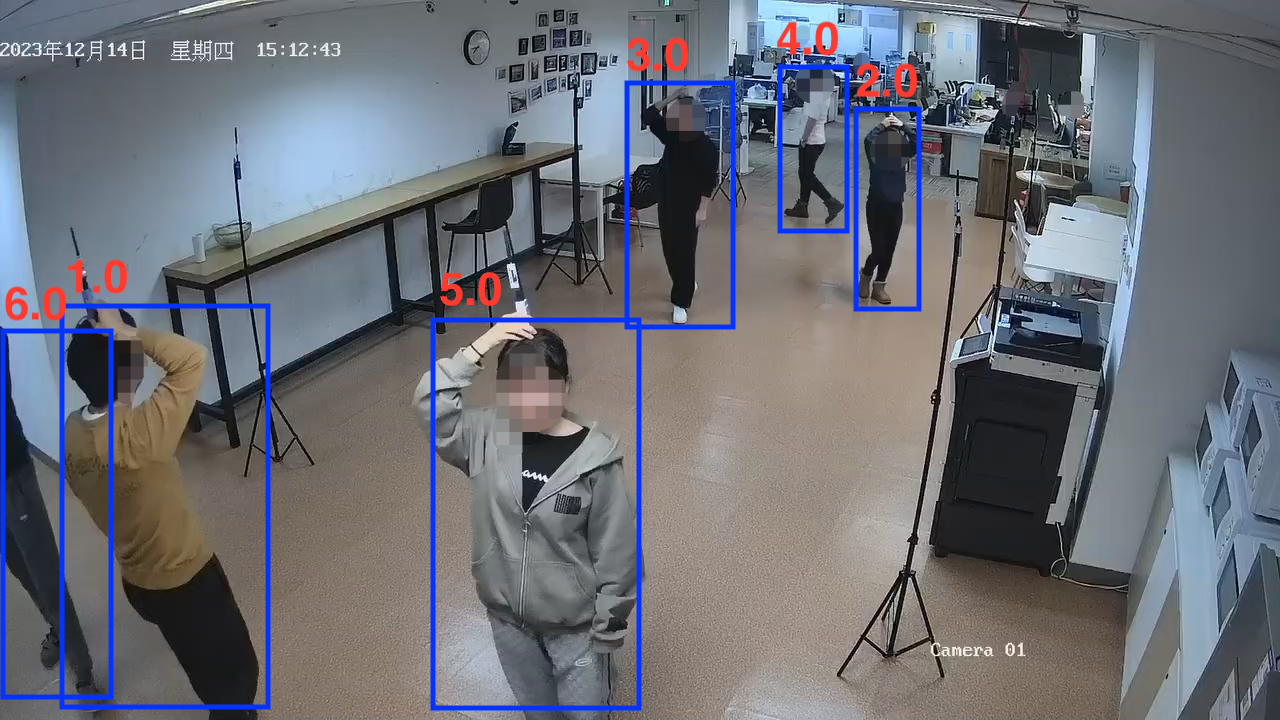}
    \caption{pixel extraction error}
    \label{sub: pixel}
  \end{subfigure}
  \hfill
  \begin{subfigure}{0.45\linewidth}
  \includegraphics[width=\linewidth]{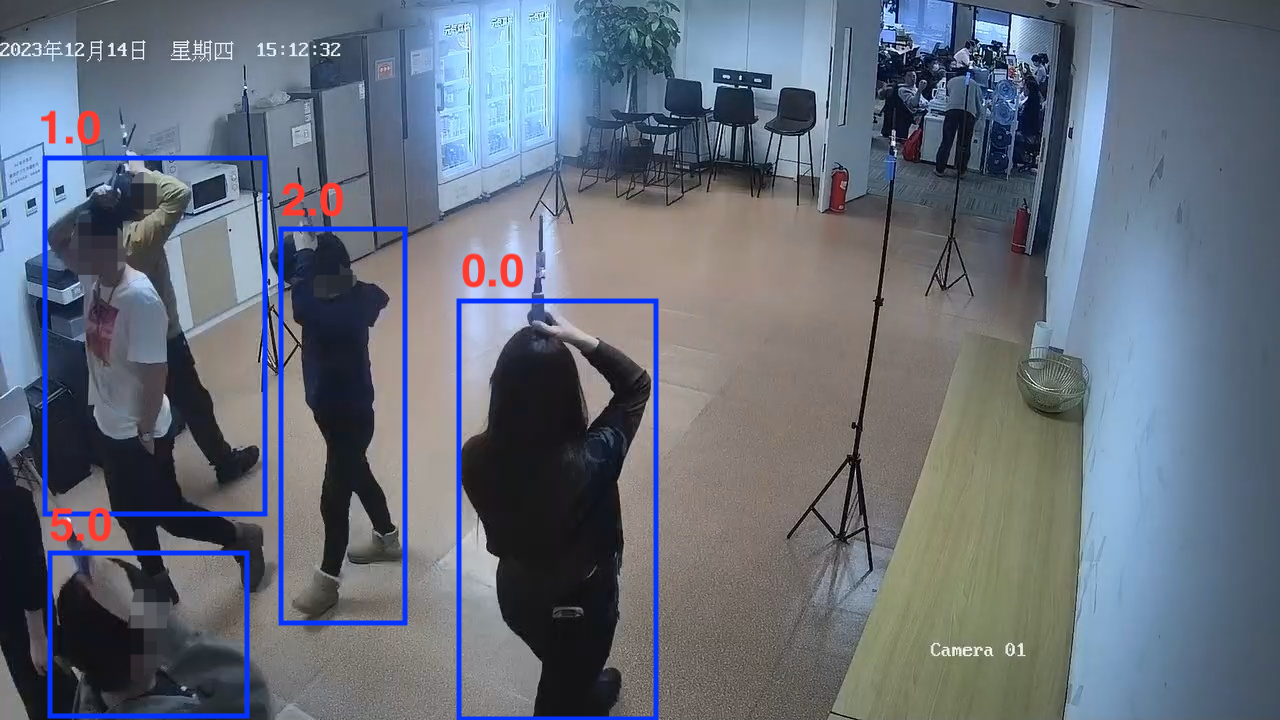}
    \caption{frequent occlusion}
    \label{sub: occlusion}
  \end{subfigure}
  
  \caption{Illustration of errors in pedestrian localization. For Figure \ref{sub: pixel}, the upper center point of detection box 6.0 deviates severely from the head pixel. For Figure \ref{sub: occlusion}, due to occlusion between objects, the detection box 1.0 wrongly contains two people.}
  \label{fig: trajectory error}
\end{figure}

Different from simulation environment and anchor localization experiments, pedestrian localization is based on detection boxes. Therefore, the precision of localization is susceptible to pixel extraction error, frequent occlusion, and incorrect identification of the detection algorithm, which pose significant challenges to accurate localization. See Figure \ref{fig: trajectory error} for illustration. Under this scenario, we check the effectiveness of anchor-based method. The pedestrian localization result across 1700 frames under different methods is shown by Table \ref{exp:real-denoising}, where the anchor-based method achieve the best performance on average distance, and can bring nearly 10 cm performance margin compared with no-anchor model for multi-camera case.

\begin{table}[htb]
    \centering
    \caption{Average distance (m) of localization methods on real-world data}
  \begin{tabular}{ccccccc}
    \toprule
    &\multicolumn{3}{c} {\textbf{Multi-Camera Case}} & \multicolumn{3}{c}{\textbf{Single-Camera Case}}\\
    \cmidrule(lr){2-4} \cmidrule(lr){5-7}
    Target ID & Init  & No Anchor & Anchor   & Init & No Anchor & Anchor  \\
    \midrule
0 & 0.635 & 0.601 & 0.510 & 0.750 & 0.767 & 0.677 \\
1 & 0.672 & 0.573 & 0.488 & 0.620 & 0.611 & 0.562 \\
2 & 0.564 & 0.552 & 0.485 & 0.623 & 0.641 & 0.627 \\
3 & 0.730 & 0.757 & 0.685 & 0.901 & 0.965 & 0.874 \\
4 & 0.641 & 0.543 & 0.453 & 0.905 & 0.871 & 0.839 \\
5 & 0.602 & 0.541 & 0.452 & 0.674 & 0.705 & 0.599 \\
6 & 0.675 & 0.595 & 0.504 & 0.895 & 0.868 & 0.761 \\
\textbf{overall} & 0.644 & 0.589 & 0.504 & 0.756 & 0.767 & 0.699 \\
    \bottomrule
  \end{tabular}
\label{exp:real-denoising}
\end{table}

% \vspace{-1mm}
\subsubsection{Trajectory Localization with Smoothness Penalty}
In Section \ref{sec: smooth}, we introduce localization method with smoothness penalty. In this experiment, We test the localization result on various batch sizes $T =1,3,5,7$, and set the weight for smoothness penalty term as $\rho=60$ (defined in \eqref{eq:rho}). In addition, to measure the stability of localization result, we add the standard deviation of localization distance as performance metric. The case where batch size equals to 1 revovers the results in Section \ref{sec:exp-pedestrian}. Experiment results on total 1700 frames are exhibited in Table \ref{tab:batch size}, from which we notice that after adding smoothness penalty, there is a significant improvement in localization accuracy under the single-camera case. Further, we observe that in both single and multi camera cases, the standard deviation of distance decreases, indicating that introducing smoothness penalty enhances the stability of anchor-based method.

% \vspace{-1mm}
\begin{table}[htb]
\centering
\caption{Average and standard deviation of distance (m) under different batch sizes.}
\begin{tabular}{  >{\centering\arraybackslash}c
  >{\centering\arraybackslash}p{2cm}
  >{\centering\arraybackslash}p{1.5cm}
  >{\centering\arraybackslash}p{0.5cm}
  >{\centering\arraybackslash}p{2cm}
  >{\centering\arraybackslash}p{1.5cm}}
\toprule
 & \multicolumn{2}{c}{\textbf{Multi-Camera Case}} &  & \multicolumn{2}{c}{\textbf{Single-Camera Case}} \\ \cline{2-3} \cline{5-6} 
Batch size & Average distance   & Distance std. &  & Average distance & Distance std. \\ \midrule
1 & 0.504 & 0.341 &  &  0.699& 0.512 \\
3 & 0.496 & 0.319 &  &  0.599&  0.412\\
5 & 0.496 & 0.317 &  &  0.601&  0.415\\
7 & 0.495 & 0.318 &  &  0.603&  0.408\\ \bottomrule
\end{tabular}
\label{tab:batch size}
\end{table}
% \vspace{-2mm}

\subsection{Experiments on Public Dataset}\label{sec:exp-public}

We perform experiments on the real-world dataset WildTrack \cite{chavdarova2018wildtrack} and synthetic dataset MultiviewX \cite{hou2020multiview}. We compare the performance of initial estimation, the no-anchor method, and the proposed anchor-based method (4, 10 anchors). For each camera, we select 10 points with annotated positions as anchors. We use the average and standard deviation of the localization distance as well as the ratio of improved cases against initial estimation as evaluation metrics. From Table \ref{tab:Q1}, anchor-based method achieves more accurate and stable results on the WildTrack dataset. The performance margin in the MultiviewX dataset is relatively small, possibly due to the synthetic nature of the proxy imaging process.

\begin{figure}[htb]
    \centering
    \includegraphics[width=0.8\linewidth]{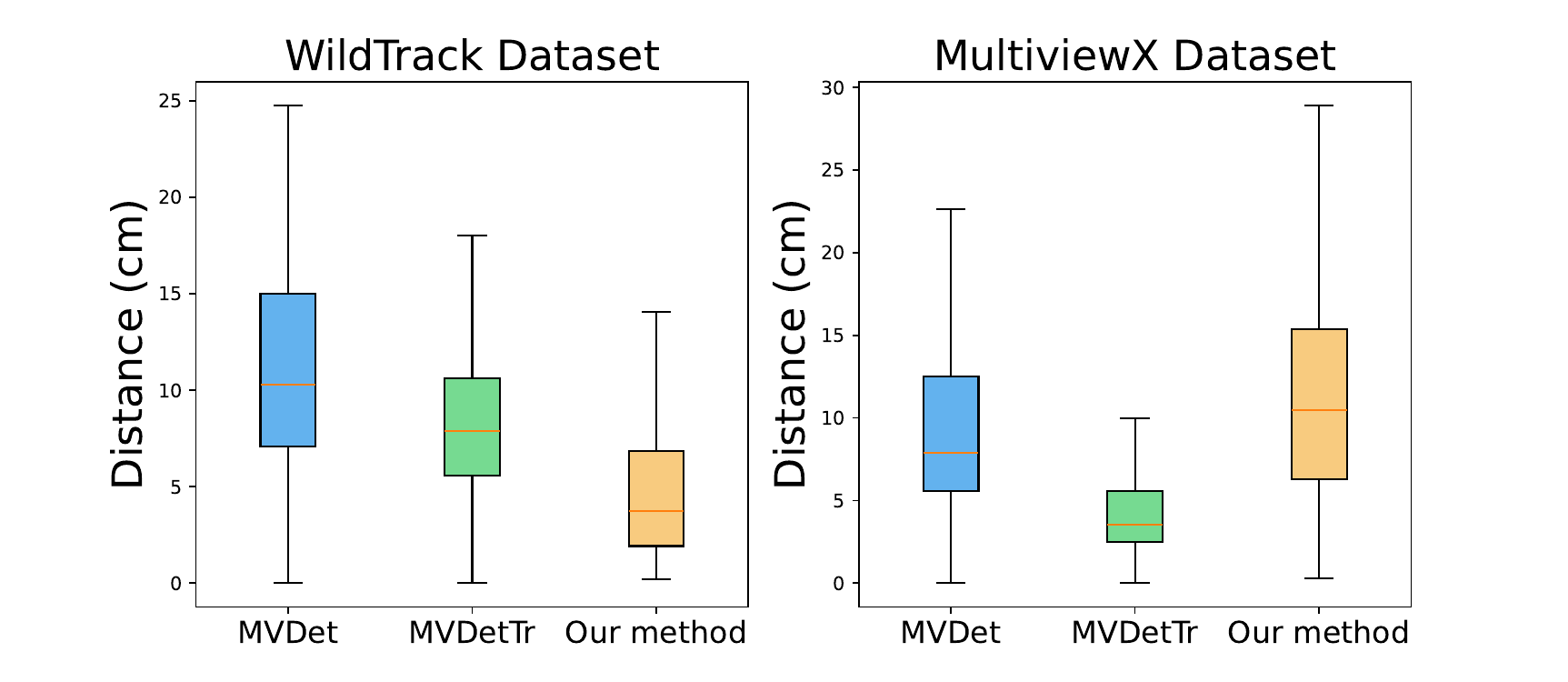}
    
    \caption{Comparison with SOTA methods \cite{hou2020multiview,hou2021multiview} on public datasets WildTrack and MultiviewX, with performance measured by the average distance (cm).}
    
    \label{fig:Q2}
\end{figure}

We also add state-of-the-art neural network-based MVDet \cite{hou2020multiview} and MVDeTr \cite{hou2021multiview} as benchmarks, which conduct joint detection and localization tasks. For fair comparison, we select accurately detected results of \cite{hou2020multiview,hou2021multiview} on test set by Hungarian algorithm, and compare the precision and stability of localization results. The results are shown in Figure \ref{fig:Q2}. Even though our method simply uses detection boxes while benchmark methods rely on whole image information, we achieve better precision on WildTrack dataset and comparable result on MultiviewX dataset. In addition, we remark that our method gains advantage on simplicity, explainability and less reliance on training data.

\begin{table}[htb]
\centering
\caption{Localization results on WildTrack and MultiviewX datasets. The average and standard deviation of distance are measured in centimeter.}

\label{tab:Q1}

\begin{tabular}{ccccc}
\toprule
 \multicolumn{1}{c}{Dataset} & \multicolumn{1}{c}{Method} & \multicolumn{1}{c}{Average Distance} & \multicolumn{1}{c}{Distance std.} & Improvement \\ 
 \midrule
\multicolumn{1}{c}{\multirow{4}{*}{WildTrack}} & Init & 13.1 & \textbf{5.78} & - \\
\multicolumn{1}{c}{} & No anchor & 9.6 & 7.99 & 76\% \\
\multicolumn{1}{c}{} & 4 anchors & \textbf{8.01} & 6.5 & 81\% \\
\multicolumn{1}{c}{} & 10 anchors & 8.09 & 6.27 & \textbf{86\%} \\ 
\midrule
\multicolumn{1}{c}{\multirow{4}{*}{MultiviewX}} & Init & 15.67 & 8.09 & - \\
\multicolumn{1}{c}{} & No anchor & 13.45 & 7.08 & 60\% \\
\multicolumn{1}{c}{} & 4 anchors & 15.14 & 9.44 & 50\% \\
\multicolumn{1}{c}{} & 10 anchors & \textbf{11.97} &\textbf{7.00} & \textbf{61\%} \\ 
\bottomrule
\end{tabular}

\end{table}

\section{Conclusion}

In this article, we investigate the multi-camera pedestrian localization problem, specifically considering scenarios where the given camera parameters may be inaccurate and re-calibration is costly. In such cases, locations determined by reprojection error minimization may deviate from the ground truth. Therefore, we propose an anchor-based localization method to reduce the errors introduced by inaccurate camera parameters. We provide a theoretical interpretation of the robustness of this method to justify its effectiveness. Additionally, we incorporate a temporal smoothness penalty to further enhance localization precision and stability. Experiments conducted on simulated, real-world, and public datasets demonstrate the effectiveness of this method in improving localization precision and robustness against noise in camera parameters.

% \begin{acknowledgements}
% If you'd like to thank anyone, place your comments here
% and remove the percent signs.
% \end{acknowledgements}

\noindent \textbf{Author Contributions} \textit{Wanyu Zhang}: Methodology, Code, Analysis, Writing--Original Draft, Visualization \textit{Jiaqi Zhang}: Code, Experimental Validation \textit{Dongdong Ge}: Supervision, Writing--Review \& Editing \textit{Yu Lin, Huiwen Yang}: Methodology, Real-world Data, Writing--Review \textit{Huikang Liu}: Analysis, Supervision, Writing--Review \& Editing \textit{Yinyu Ye}: Methodology, Supervision, Writing--Review \& Editing

\noindent \textbf{Funding} The authors declare that no funds, grants, or other support were received during the preparation of this manuscript.

\noindent \textbf{Code availability} All data and code are publicly available via \href{https://github.com/zwyhahaha/AnchorLocalization.git}{AnchorLocalization}.

\section*{Declarations}

\textbf{Conflict of interest} The authors declare that they have no conflict of interest.

\noindent \textbf{Ethics approval and consent to participate} The authors consent to participate in this study.

\noindent \textbf{Consent for publication} The authors give our consent for publication of this manuscript in the journal of Optimization and Engineering.

% \clearpage  

\appendix

\section{Supplementary Experiments on Simulated Data}

\subsection{Robustness to Combinatorial Camera Parameter Errors} \label{app:exp-simu}

To test the robustness of the proposed method against combinatorial camera parameter errors, we select various combinations of perturbation levels, with the results presented in Table \ref{tab:combination of errors}. Under these different parameter combinations, the anchor-based method significantly reduces the average distance to the ground-truth position. As the level of noise increases, the performance gap between the no-anchor method and the anchor-based method becomes more pronounced. For better benchmarking, we evaluate the no-anchor method using the ground-truth camera parameters, which represent the best achievable results. An increasing number of anchors further nudges the localization precision closer to the best possible results.

\begin{table}[htb]
\centering
\caption{Average distance (m) under different combinations of errors. $N_a$ represents the number of anchors.}
\begin{tabular}{ccccccccc}
\toprule
\textbf{Rx} & \textbf{Ry} & \textbf{T} & \textbf{D} & \textbf{} & \textbf{$N_a=0$} & \textbf{$N_a=4$} & \textbf{$N_a=8$} & Ground Truth \\ \cline{1-4} \cline{6-9} 
0.25 & 0.25 & 0.05 & 0.25 &  & 0.095 & 0.066 & 0.060 & 0.018 \\ 
0.25 & 0.25 & 0.05 & 0.50 &  & 0.163 & 0.135 & 0.110 & 0.018 \\ 
0.25 & 0.25 & 0.10 & 0.25 &  & 0.099 & 0.081 & 0.065 & 0.018 \\ 
0.25 & 0.25 & 0.10 & 0.50 &  & 0.205 & 0.185 & 0.134 & 0.018 \\ 
0.25 & 0.50 & 0.05 & 0.25 &  & 0.131 & 0.085 & 0.087 & 0.018 \\ 
0.25 & 0.50 & 0.05 & 0.50 &  & 0.181 & 0.154 & 0.119 & 0.018 \\ 
0.25 & 0.50 & 0.10 & 0.25 &  & 0.162 & 0.101 & 0.082 & 0.018 \\ 
0.25 & 0.50 & 0.10 & 0.50 &  & 0.233 & 0.141 & 0.122 & 0.018 \\ 
0.50 & 0.25 & 0.05 & 0.25 &  & 0.090 & 0.078 & 0.061 & 0.018 \\ 
0.50 & 0.25 & 0.05 & 0.50 &  & 0.159 & 0.121 & 0.107 & 0.018 \\ 
0.50 & 0.25 & 0.10 & 0.25 &  & 0.165 & 0.080 & 0.073 & 0.018 \\ 
0.50 & 0.25 & 0.10 & 0.50 &  & 0.236 & 0.162 & 0.147 & 0.018 \\ 
0.50 & 0.50 & 0.05 & 0.25 &  & 0.120 & 0.080 & 0.070 & 0.020 \\ 
0.50 & 0.50 & 0.05 & 0.50 &  & 0.210 & 0.180 & 0.160 & 0.020 \\ 
0.50 & 0.50 & 0.10 & 0.25 &  & 0.160 & 0.090 & 0.090 & 0.020 \\ 
0.50 & 0.50 & 0.10 & 0.50 &  & 0.220 & 0.190 & 0.140 & 0.020 \\ 
\bottomrule
\end{tabular}
\label{tab:combination of errors}
\end{table}

\subsection{Robustness to Pixel Extraction Errors} \label{app:exp-pix}

\begin{figure}[htb]
    \centering
    \includegraphics[width=0.8\linewidth]{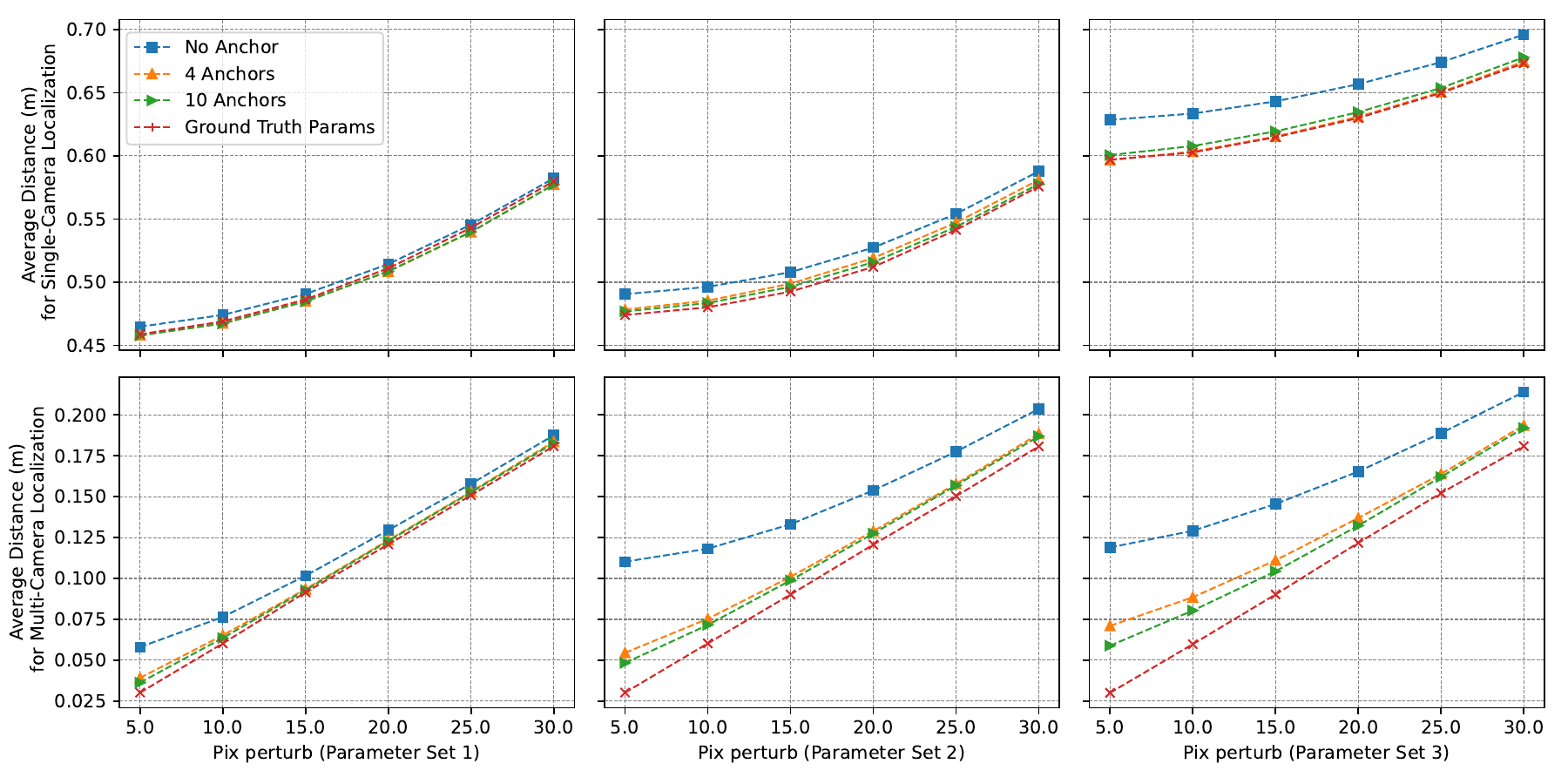}
    \caption{Average distance (m) of different localization methods under different perturbation levels of observed pixels. For camera parameter error setting, \textit{left (Set 1):} Rx error=Ry error=0.2, T error=D error=0.05. \textit{Center (Set 2):} Rx error=Ry error=0.2, T error=D error=0.1. \textit{Right (Set 3):} Rx error=Ry error=0.4, T error=D error=0.1.}
    \label{fig:simu-pix-err-loc}
\end{figure}

\begin{figure}[htb]
    \centering
    \includegraphics[width=0.8\linewidth]{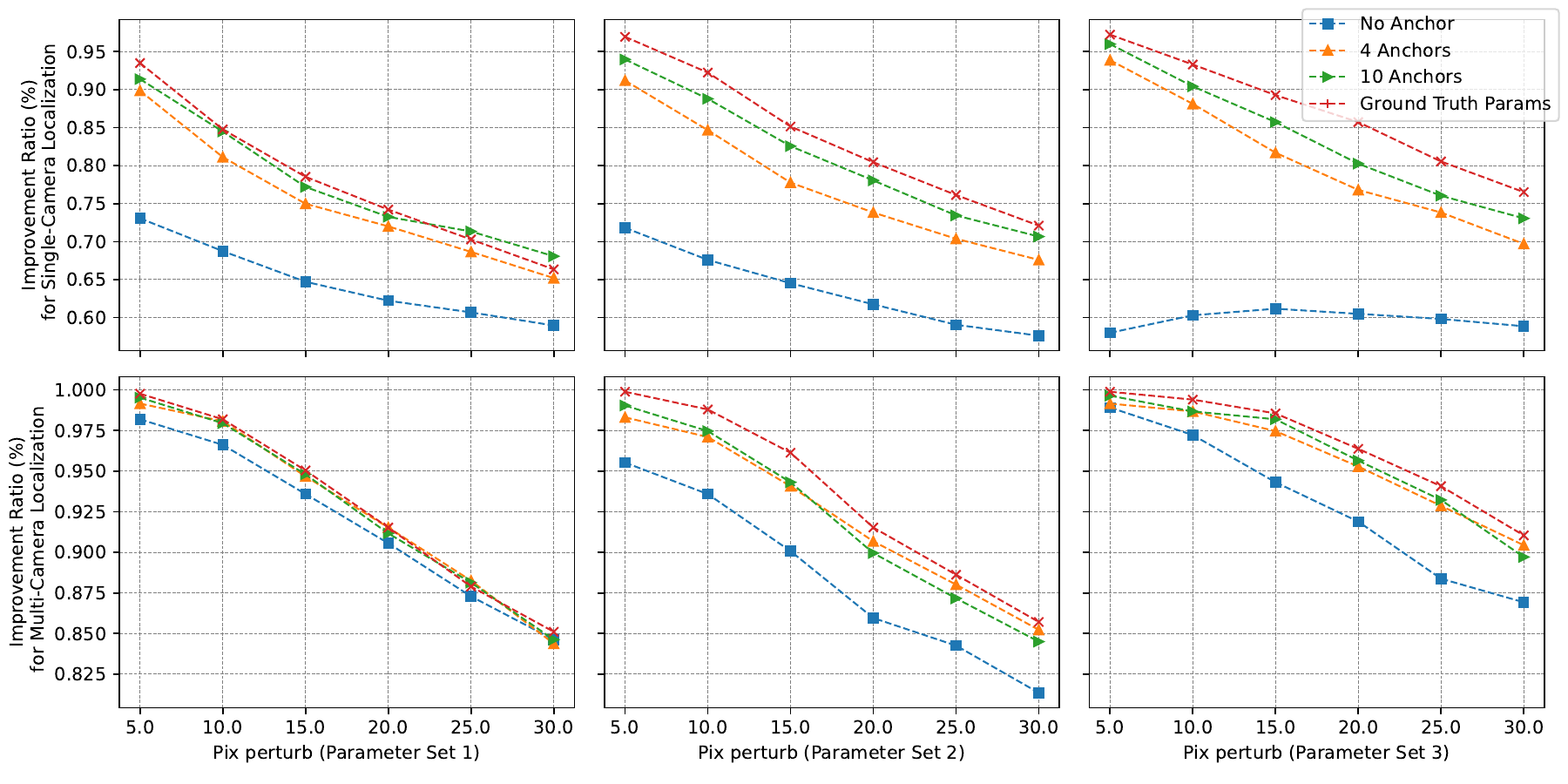}
    \caption{Improvement Ratio of different localization methods under different perturbation levels of observed pixels. Camera parameter error settings are the same as Figure \ref{fig:simu-pix-err-loc}}
    \label{fig:simu-pix-err-loc-imp}
\end{figure}

We introduce pixel extraction errors to simulate scenarios in which an inaccurate pixel substitutes the target's actual observed pixel. For example, when the upper center point of a detection box is used to represent the head point in an image, a pixel extraction error is incurred. Such errors are pronounced in real-world settings. We fix 3 sets of erroneous camera parameters, and vary pixel perturbation level $\delta$ from 5 to 30 pixels. The experiment result is shown in Figure \ref{fig:simu-pix-err-loc} and Figure \ref{fig:simu-pix-err-loc-imp}. Across different levels of pixel perturbation, the anchor-based method achieves a lower average distance and a higher improvement ratio compared to the no-anchor method.

\section{Supplementary Experiments on Real-World Data}

\subsection{Representative Point Selection} 
\label{sec:exp-rep}

Unlike the simulation and anchor localization experiments, where the target is represented as a "point," real-world pedestrian localization involves the problem of representative point selection, with potential choices including the ankle and foot points. For the head point, we treat the upper center point of the detection box as the head pixel and subsequently determine the 3D position of the head point. For the ankle point, we select the bottom center point and identify the grounding point that minimizes the reprojection errors.

\begin{figure}[htb]
  \centering
  \begin{subfigure}{0.45\linewidth}
    \includegraphics[width=\linewidth]{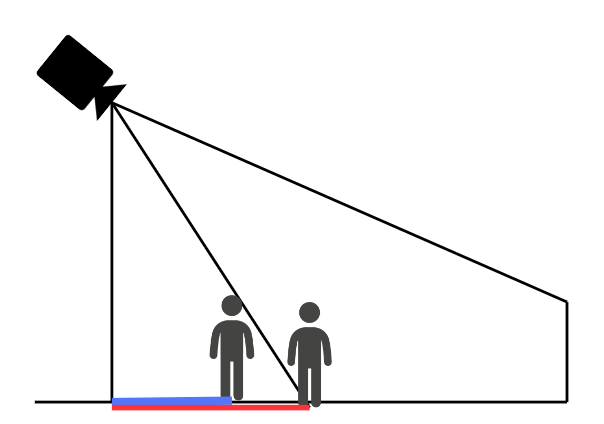}
    % \fbox{\rule{0pt}{0.5in} \rule{.9\linewidth}{0pt}}
    \caption{Blind View}
  \end{subfigure}
  \hfill
  \begin{subfigure}{0.45\linewidth}
  \includegraphics[width=\linewidth]{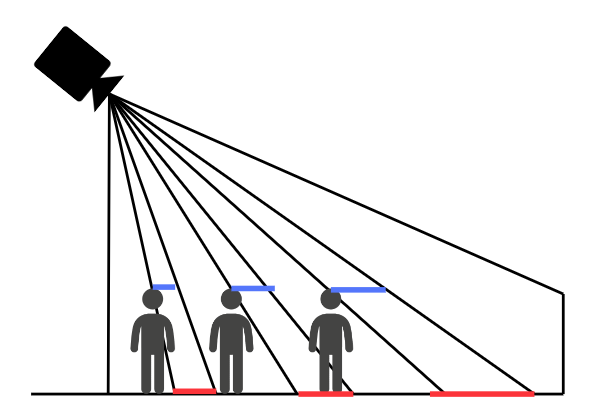}
    \caption{Sensitivity to errors}
  \end{subfigure}
  \caption{Comparison of using head and ankle point. \emph{Left}: given the FOV of a camera, the blue (red) line represents the blind view of head (ankle) point, respectively. \emph{Right}: given the same angular error, the blue (red) lines represent the errors of head (ankle) points, respectively.}
  \label{fig:cmp-ankle-head}
\end{figure}

In this section, we conduct ablation experiments to investigate the influence of representative point selection problem. The experiment result is shown by Table \ref{tab: head ankle}. For both the no-anchor method and the anchor-based methods, selecting the head as the representative point achieves higher localization precision compared to the ankle point. We identify several reasons for the advantages of using head points. In real-world images, ankle points are often occluded. Additionally, compared to head points, ankle points are more susceptible to: (a) larger blind view, (b) increased sensitivity to camera parameter errors. See Figure \ref{fig:cmp-ankle-head} for illustration.

\begin{table}[htb]
\centering
\caption{Average distance (m) of using ankle and head as representative point.}
\begin{tabular}{
  >{\centering\arraybackslash}c
  >{\centering\arraybackslash}p{1.3cm}
  >{\centering\arraybackslash}p{1.6cm}
  >{\centering\arraybackslash}p{1.6cm}
  >{\centering\arraybackslash}p{1.6cm}
  >{\centering\arraybackslash}p{1.6cm}}
\toprule
& &\multicolumn{2}{c} {\textbf{No Anchor}} & \multicolumn{2}{c}{\textbf{Anchor-based}}\\
\cmidrule(lr){3-4} \cmidrule(lr){5-6}
    Target ID & Init & Ankle & Head & Ankle & Head  \\
\midrule
0 & 0.655 & 0.656 & 0.630 & 0.587 & 0.540 \\
1 & 0.667 & 0.670 & 0.577 & 0.607 & 0.495 \\
2 & 0.576 & 0.622 & 0.572 & 0.566 & 0.515 \\
3 & 0.759 & 0.844 & 0.793 & 0.717 & 0.717 \\
4 & 0.684 & 0.709 & 0.596 & 0.591 & 0.515 \\
5 & 0.614 & 0.648 & 0.569 & 0.546 & 0.477 \\
6 & 0.699 & 0.714 & 0.625 & 0.617 & 0.532 \\
\textbf{overall} & 0.662 & 0.689 & 0.616 & 0.600 & 0.535\\
\bottomrule
\end{tabular}
\label{tab: head ankle}
\end{table}

% \clearpage

\bibliographystyle{splncs04}
\bibliography{reference}

\end{document}